\documentclass[a4paper, 10pt, conference]{ieeeconf}      % Use this line for a4 paper

\IEEEoverridecommandlockouts                              % This command is only needed if
\usepackage[pdftex]{graphicx} % for pdf, bitmapped graphics files
\graphicspath{{img/pdf/}{img/png/}{img/jpg/}{img/eps/}{img/svg/}}
\DeclareGraphicsExtensions{.pdf,.png,.jpg,.eps}
\usepackage[caption=false,font=footnotesize]{subfig}
\usepackage{stfloats}
\usepackage{amsmath} % assumes amsmath package installed
\usepackage{amssymb}  % assumes amsmath package installed
\usepackage{siunitx}
\usepackage{hyperref}
\usepackage{booktabs}

\usepackage{bm}

\title{\LARGE \bf
A Modular Bio-inspired Robotic Hand with High Sensitivity
}

\author{Chao Liu, Andrea Moncada, Hanna Matusik, Deniz Irem Erus, and Daniela Rus% <-this % stops a space
\thanks{*This work was supported by the Gwangju Institute of Science and Technology (GIST)}% <-this % stops a space
\thanks{The authors are with the Computer Science \& Artificial
  Intelligence Laboratory, Massachusetts Institute of Technology, 32
  Vassar St, Cambridge, MA 02142, USA {\tt\small
    chaoliu@csail.mit.edu, amoncada@mit.edu, hmatusik@mit.edu,
    dierus@mit.edu, rus@csail.mit.edu}}%
}

\begin{document}

\maketitle
\thispagestyle{empty}
\pagestyle{empty}

%%%%%%%%%%%%%%%%%%%%%%%%%%%%%%%%%%%%%%%%%%%%%%%%%%%%%%%%%%%%%%%%%%%%%%%%%%%%%%%%
\begin{abstract}
  While parallel grippers and multi-fingered robotic hands are well
  developed and commonly used in structured settings, it remains a
  challenge in robotics to design a highly articulated robotic hand
  that can be comparable to human hands to handle various daily
  manipulation and grasping tasks. Dexterity usually requires more
  actuators but also leads to a more sophisticated mechanism design
  and is more expensive to fabricate and maintain. Soft materials are
  able to provide compliance and safety when interacting with the
  physical world but are hard to model. This work presents a hybrid
  bio-inspired robotic hand that combines soft matters and rigid
  elements. Sensing is integrated into the rigid bodies resulting in a
  simple way for pose estimation with high sensitivity. The proposed
  hand is in a modular structure allowing for rapid fabrication and
  programming. The fabrication process is carefully designed so that a
  full hand can be made with low-cost materials and assembled in an
  efficient manner. We demonstrate the dexterity of the hand by
  successfully performing human grasp types.
\end{abstract}

%%%%%%%%%%%%%%%%%%%%%%%%%%%%%%%%%%%%%%%%%%%%%%%%%%%%%%%%%%%%%%%%%%%%%%%%%%%%%%%%
\section{INTRODUCTION}

Dexterous manipulation and grasping have been and are still being
actively explored in robotics, including motion planning in high
dimensional space~\cite{Sucan-ompl-ram-2012}, finger contact
model~\cite{Salisbury-finger-contact-model-1983}, in-hand manipulation
with visual
feedback~\cite{Andrychowicz-RL-inhand-manipulation-ijrr-2020} or
tactile sensing
feedback~\cite{Hoof-RL-inhand-manipulation-tactile-humanoids-2015}. While
it is a common daily task for humans to interact with various objects,
manipulation and grasping activities in the real world are still
challenging for robots. In order to successfully deploy a robot that
can handle manipulation and grasping tasks in daily life, an
intelligent system that is capable of adapting to various scenarios
and utilizing different strategies is needed. Robotic hardware is
fundamental for achieving this goal.

In order to handle more general manipulation tasks, more complex robot
hand designs are usually required. More degrees of freedom (DoFs) are
helpful to address various scenarios but may not be practical in real
applications. On the other hand, simple hands are often more robust
and can usually offer more reliable performance but have limited
applications. For example, grippers are commonly deployed in
industry. Vacuum grippers are easy to use for moving packages and are
also suitable for picking up and placing small components. Parallel
jaws with single actuators are also popular, e.g. Willow Garage Velo
2G~\cite{velo-hand}. Simplified multi-fingered hands could offer more
dexterity. These hands are simple and nonanthropomorphic leading to
low cost. However, they usually can perform a restricted class of
tasks and cannot achieve generality. Highly articulated robotic hands
are promising for handling a much wider range of tasks. Many
researchers have been trying to replicate human hands by using more
sophisticated mechanism designs and adding more actuators and sensors,
e.g. Shadow robotic hand~\cite{Kochan-shadow-hand-2005}. However,
these hands are expensive and difficult to fabricate, calibrate, and
maintain. One solution to reduce the complexity and also maintain the
anthropomorphic design is to use underactuation, namely using fewer
number of actuators to achieve more DoFs. For these designs,
cable-driven approaches have been commonly used.

Compliance and safety are critical for robots when interacting with
the physical world, especially with human beings. Using soft materials
is a promising solution to make compliant robots. Such robots could be
more adaptive to different tasks without relying on sophisticated
control strategies. Moreover, soft materials have greater interaction
safety by absorbing impact energy. However, soft robots usually
require heavy actuation systems (pneumatic or hydraulic), and are also
difficult to model due to the lack of precision. Because of the design
and the fabrication techniques of soft
robots~\cite{Rus-soft-robot-review-nature-2015}, soft robotic hands
are usually limited to several single-DoF fingers and are not capable
of doing dexterous in-hand manipulation tasks compared with rigid
robotic hands.

In nature, it is common that soft materials and rigid bodies are
combined together to construct tissues. This combination can create
strong, precise, and also compliant biological systems. This fact has
inspired the robotics community to explore soft-rigid structures. In
this paper, we show that the combination of soft materials and rigid
elements can generate novel robotic tissues that are able to inherit
the benefits of both materials leading to performance that cannot be
achieved by either one independently. The topological connection of
the rigid elements defines the kinematic structure. The electronics
can be integrated for sensing and measurement. A multi-layer silicone
casting process is presented in order to mix the soft materials with
the rigid elements while also maintaining the structural architecture
of these rigid elements. The resulting robotic tissue integrates soft
materials for compliance and shaping with an articulated rigid
skeleton for structural strength and pose measurement. Using this
approach, we created a bio-inspired robotic hand. We precisely cloned
the shape of a human hand and all the bones. We converted the hand
skeleton into several carefully designed bone-style linkage
systems. These linkage systems are cable-driven and have integrated
position sensors. The finger design has three joints and can be easily
customized for different number of actuators. One significant
advantage of this solution is low cost. Low-cost sensors and devices
are important for rapid prototyping and research exploration that can
provide easy customization and facilitate robotics benchmarks. For
example, low-cost position sensors using paints can be easily
customized to fit into highly space constrained modular
robots~\cite{Liu-paintpot-sensing-jmr-2021}. Furthermore, low-cost
robot platforms made by servo motors and simple mechanical components
can be easily set up for
benchmarks~\cite{Ahn-robotics-benchmark-low-cost-hardware-corl-2019}. The
robot tissue can be made using 3D printed components and low-cost
materials (cables, pins, tubes, silicone, and magnets) easily by
following our fabrication procedures. The modular design of the
robotic hand also allows customization to generate robotic grippers in
various morphologies, including the human hand
morphology. Furthermore, after mounting all five fingers on a palm, we
can derive a robotic hand that is highly similar to a human hand and
is able to generate more natural gestures rather than simple motions,
such as waving arms~\cite{Specian-quori-human-robot-tro-2021}. This
capability could benefit human-robot interaction with better
performance.

% low cost can facilitate robotics benchmark
% HRI application

\section{RELATED WORK}
\label{sec:related}

Grippers usually have simple designs and have been deployed in many
scenarios. In addition to common parallel grippers, some specially
designed grippers with more complicated cable-driven mechanisms have
shown with more capability, such as the Barrett
Hand~\cite{Ulrich-upenn-barrett-hand-icra-1988}, the
iRobot-Harvard-Yale Hand~\cite{Odhner-cable-driven-hand-ijrr-2014},
the M$^2$ Gripper~\cite{Ma-cable-driven-gripper-2015}, and an
underactuated gripper that can grasp objects
sequentially~\cite{Mucchiani-cable-underactuated-planar-gripper-icra-2020}. These
grippers are aiming for a set of tasks rather than general purpose
use.

In order to derive human-level dexterity, the human hand has been the
design objective for general-purpose robotic hands. The Utah/MIT hand
was first presented in~\cite{Jacobsen-utah-mit-hand-icra-1986} to
facilitate machine dexterity research. The joints of human hands were
analyzed that inspired the design of the DLR
hand~\cite{Grebenstein-cable-driven-finger-iros-2010}. Due to the
complexity of the human hand structure, underactuation is commonly
used and the joint motions are usually coupled. Many robotic hands
make use of cable-driven mechanisms with incorporated pulleys,
e.g.~\cite{Kuo-cable-driven-finger-icra-2015,Santina-cable-driven-hand-tro-2018,
  You-cable-driven-hand-2018}, or
springs~\cite{Min-cable-driven-anthropomorphic-hand-ral-2021} in which
actuators and electronics can be installed inside a forearm. However, carefully designed structures are needed for cable
routing. The linkage-driven approach is an
alternative~\cite{Kim-linkage-robotic-hand-nature-com-2021} but is
more difficult to achieve larger workspace. To avoid sophisticated
mechanism designs, some robotic hands are driven directly or through a
gear or a timing
pulley~\cite{Liu-dexterous-hand-iros-2008,Lee-kitchen-robotic-hand-tmech-2016},
but these hands are usually much larger.

Compliance is important for providing more adaptive grasping
capability -- this property is innate for soft robots. The PneuNet
technique~\cite{Ilievski-pneunet-actuator-2011} is commonly used for
developing soft fingers,
e.g.~\cite{Deimel-pneumatic-soft-actuator-icra-2013,Polygerinos-pneumatic-hand-rehabilitation-iros-2013,Deimel-soft-hand-ijrr-2015},
in which fingers are driven by pneumatic control systems. More DoFs
can be added by increasing the complexity of the molding structures to
contain more chambers, such
as~\cite{Zhou-pneumatic-soft-hand-ral-2018,Truby-tactile-soft-finger-robosoft-2019,Hashemi-bone-inspired-soft-hand-soro-2020}. Other
soft materials and foams have also been explored to construct soft
robotic fingers~\cite{Bauer-foam-hand-2019}.

Inspired by human hands, various skeleton structures have been
developed. Finger bones can be cast or 3D printed, and then they can be
connected with silicone~\cite{Faudzi-soft-muscle-hand-ral-2018} or
rubber
bands~\cite{Tian-soft-hand-with-skeleton-2021,Bern-soft-hand-icra-2022}. A
passive skeleton structure is made by a multimaterial 3D printing
process~\cite{Hughes-soft-skeleton-hand-scirobotics-2018} and the
joint stiffness can be controlled by jamming
particles~\cite{Gilday-soft-hand-jamming-joints-robosoft-2021}. These
approaches require complex fabrication process and do not include
sensing capabilities.

In this work, we propose a novel way to combine soft materials and
rigid elements to create bio-inspired robotic tissues. The articulated
rigid skeleton defines the kinematic structure and electronics are
embedded. The silicone casting process determines the shape of the
tissue that can be highly customized and also provide compliance. We
demonstrate this method by designing and fabricating a low-cost
human-like soft robotic hand. Compared with other soft robotic hands,
the design can be easily customized, the fabrication process is simple
and efficient, and the hand pose can be precisely measured in
real time and further visualized using the MANO
model~\cite{Romero-mano-hand-modeling-2017}.

\begin{figure}[b]
  \centering
  \subfloat[]{\includegraphics[width=0.45\textwidth]{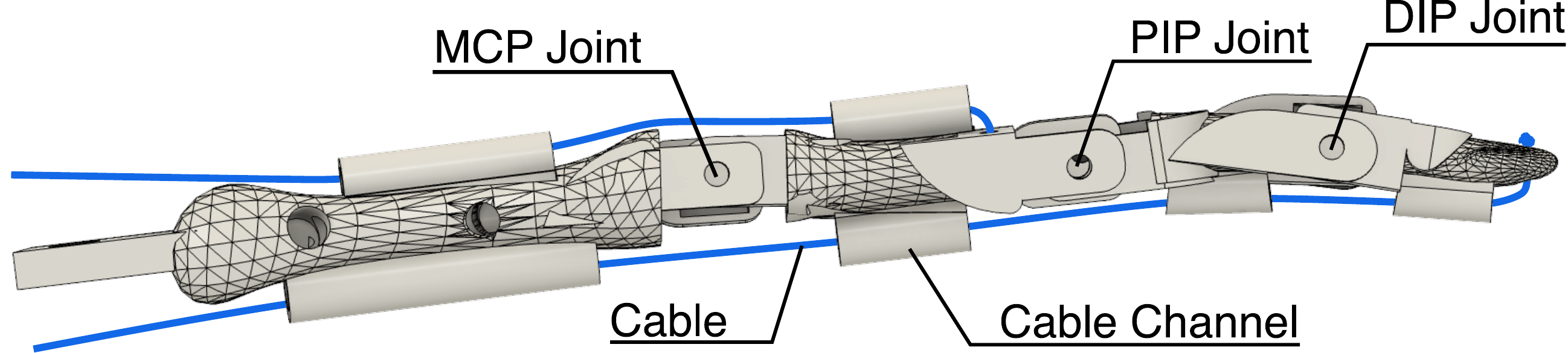}\label{fig:index-finger-bones}}
  \hfil
  \subfloat[]{\includegraphics[height=0.06\textwidth]{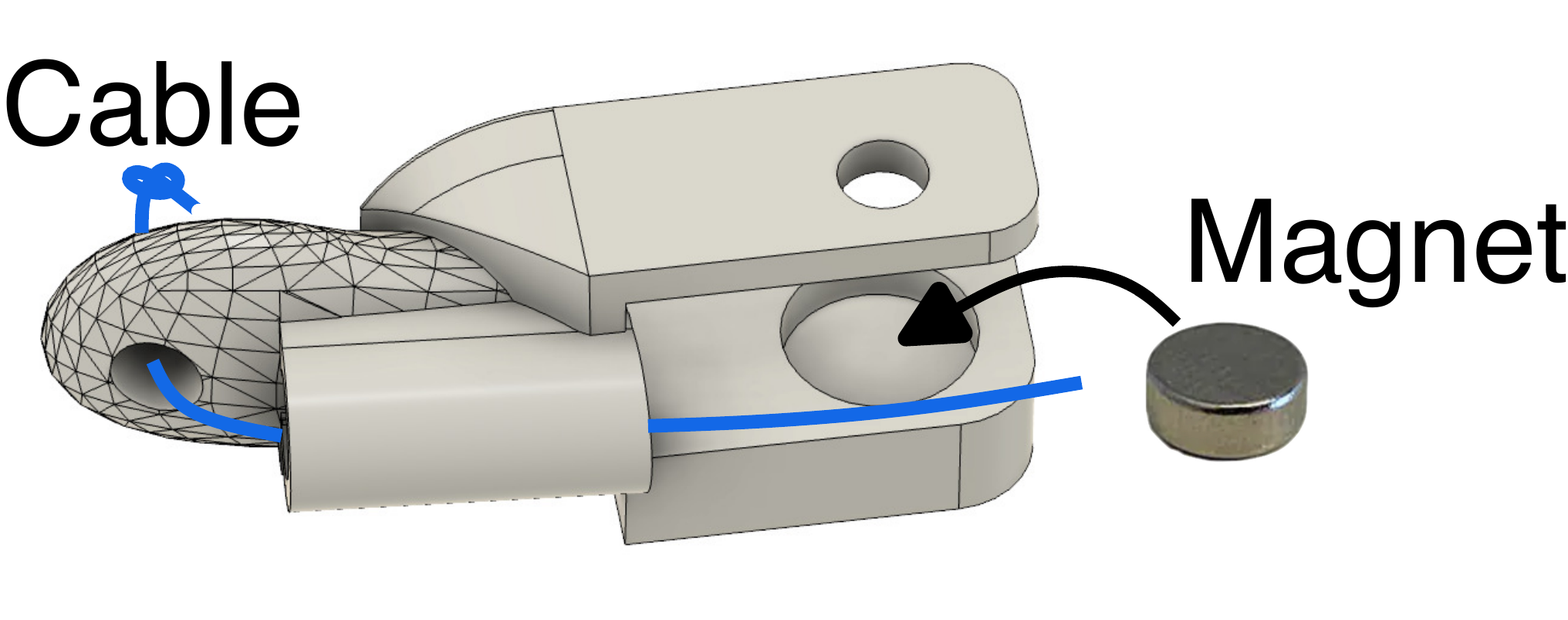}\label{fig:bone-magnet}}
  \hfil
  \subfloat[]{\includegraphics[height=0.06\textwidth]{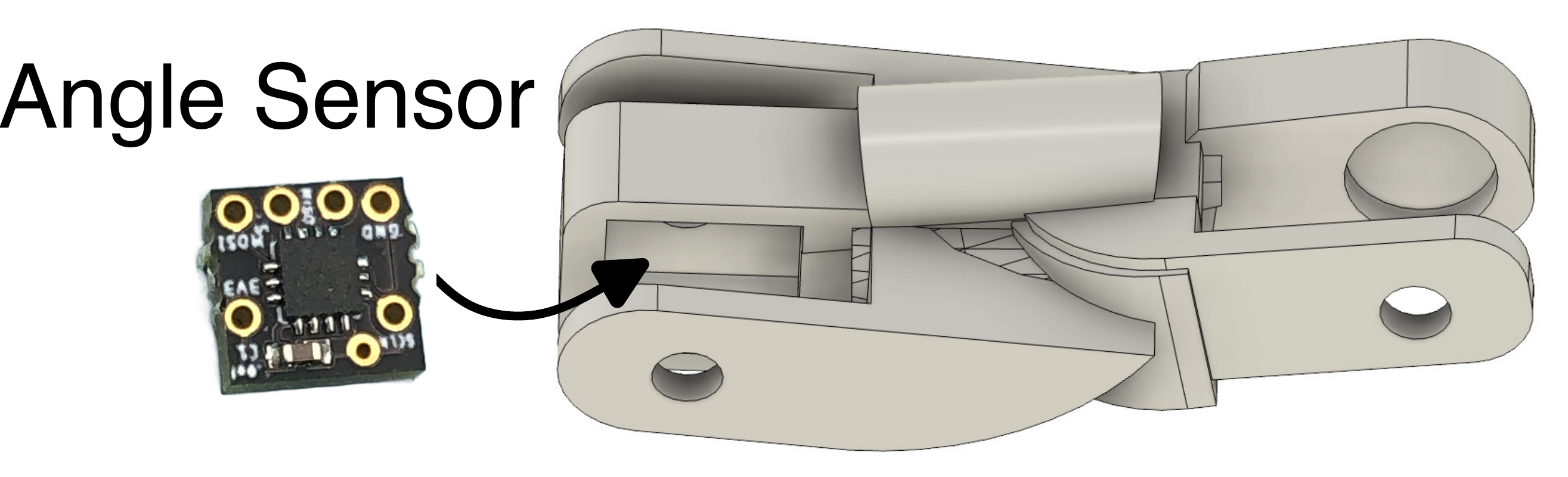}\label{fig:bone-sensor}}
  \caption{(a) The design of the bones for the index finger is based
    on the human index finger bones. There are four bones connected
    with three pin joints. (b) A diametrically magnetized magnet is placed
    inside the joint of the distal phalange. (c) An angle sensor board
    is designed and installed inside one joint of the middle phalange.}
\end{figure}

\begin{figure}[b!]
  \centering
  \includegraphics[width=0.4\textwidth,angle=10]{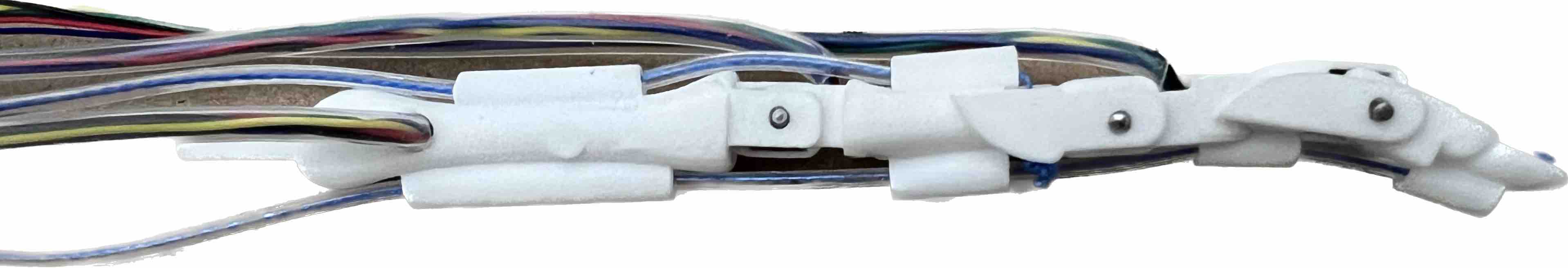}
  \caption{Fully assembled index finger skeleton with electric wires
    for angle sensors.}
  \label{fig:skeleton-assembly}
\end{figure}

 \begin{figure*}[b!]
   \centering
   \subfloat[]{\includegraphics[width=0.2\textwidth]{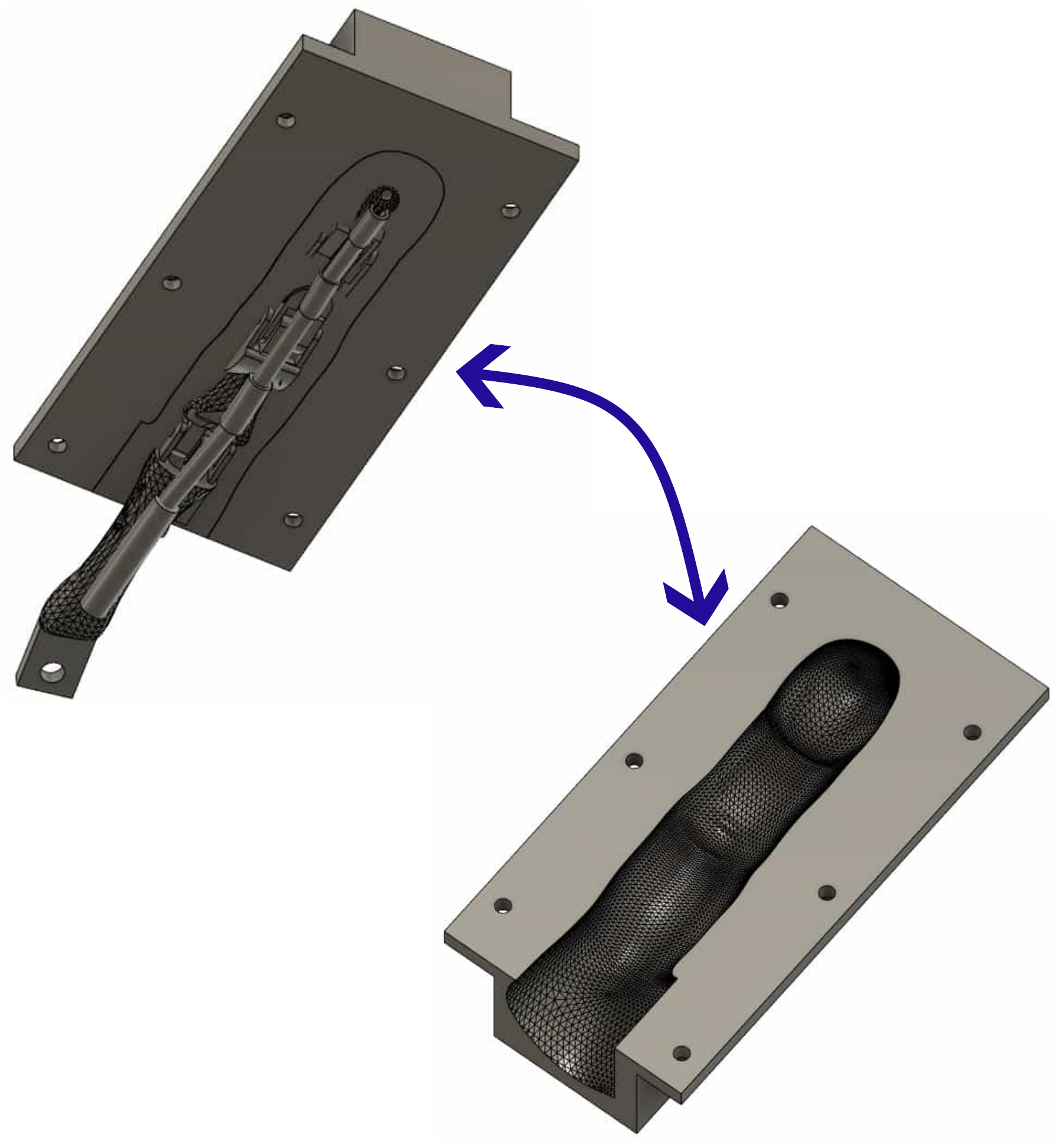}\label{fig:mold-step1}}
   \hfil
   \subfloat[]{\includegraphics[width=0.23\textwidth]{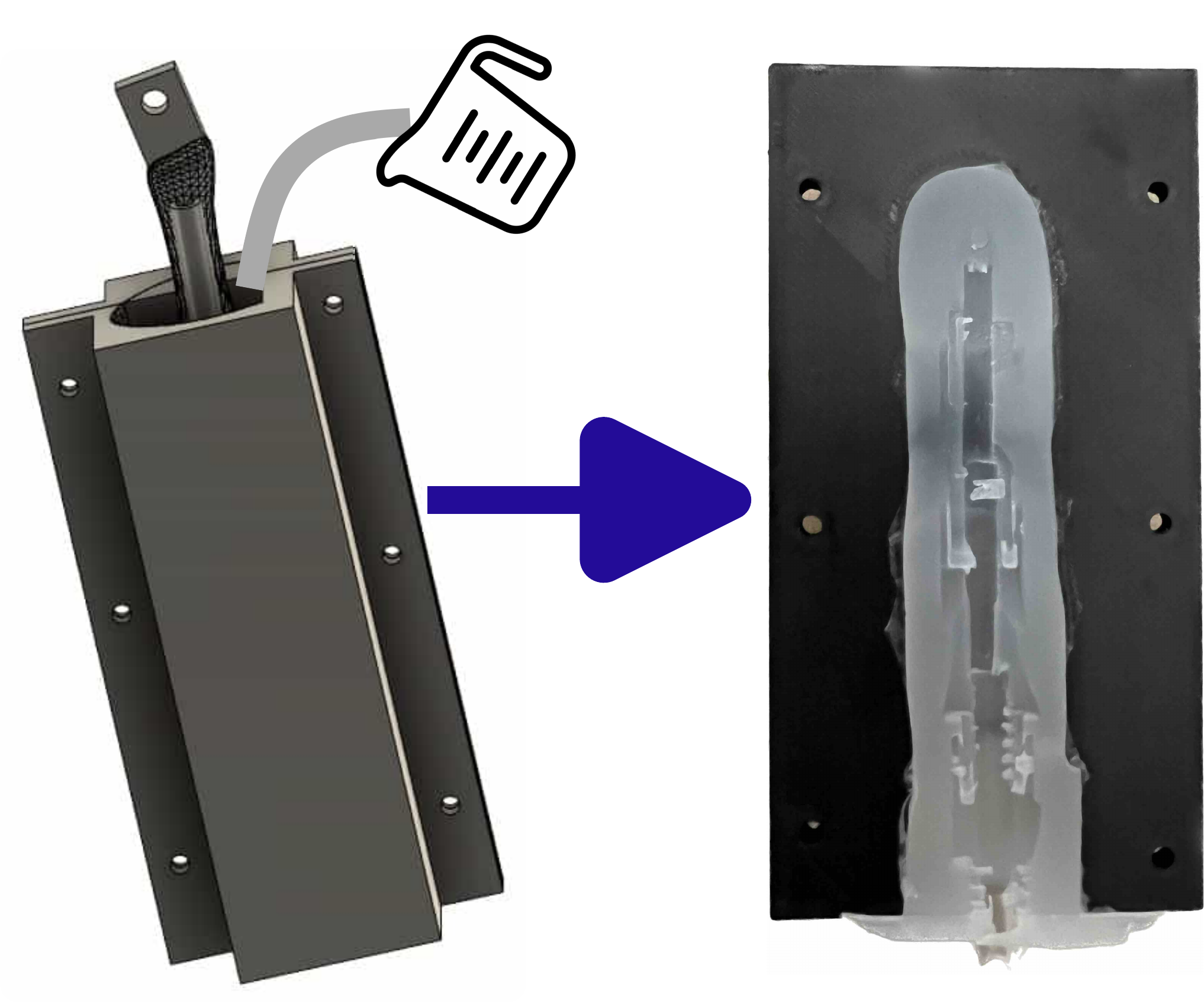}\label{fig:mold-step2}}
   \hfil
   \subfloat[]{\includegraphics[width=0.2\textwidth]{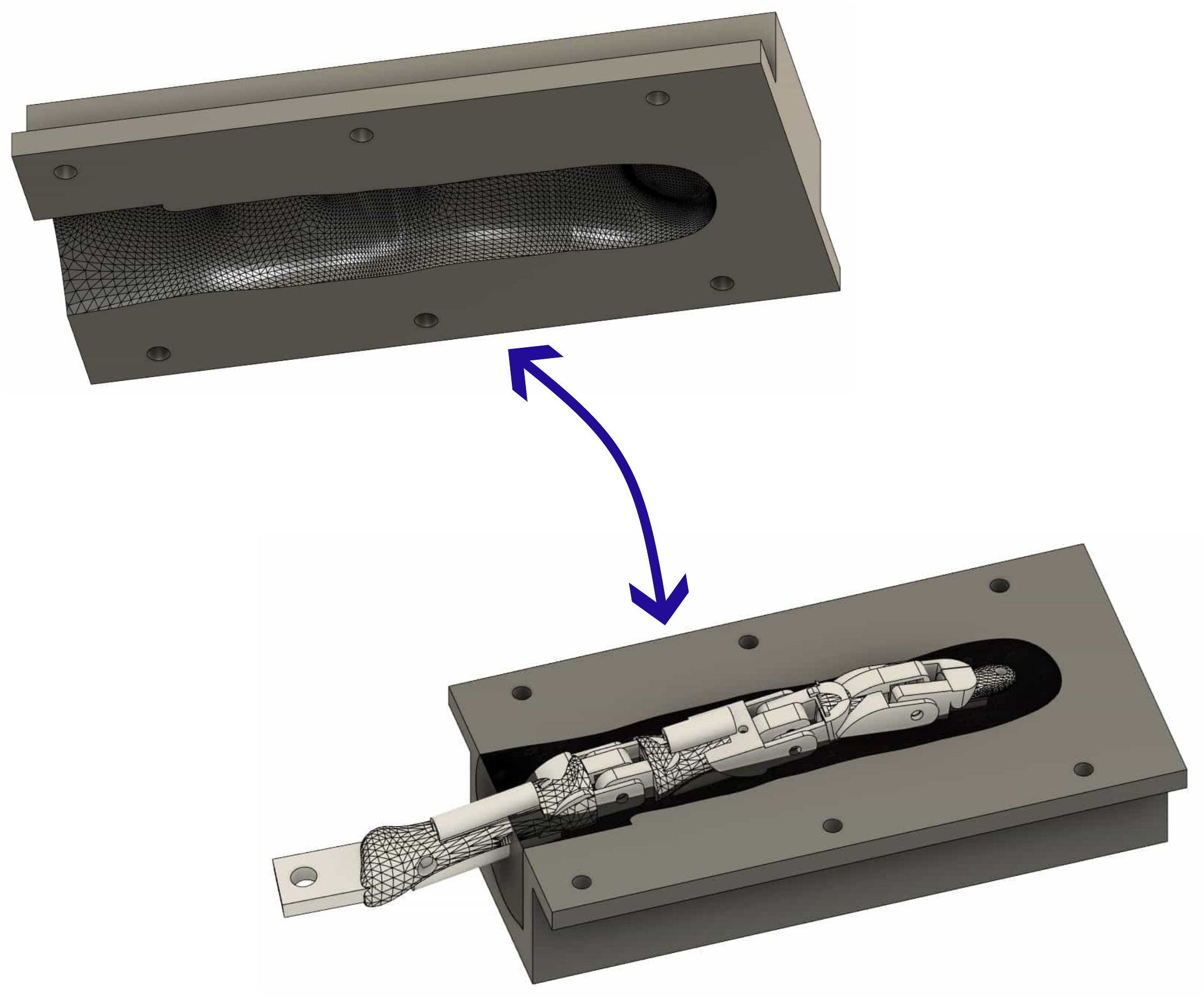}\label{fig:mold-step3}}
   \hfil
   \subfloat[]{\includegraphics[width=0.2\textwidth]{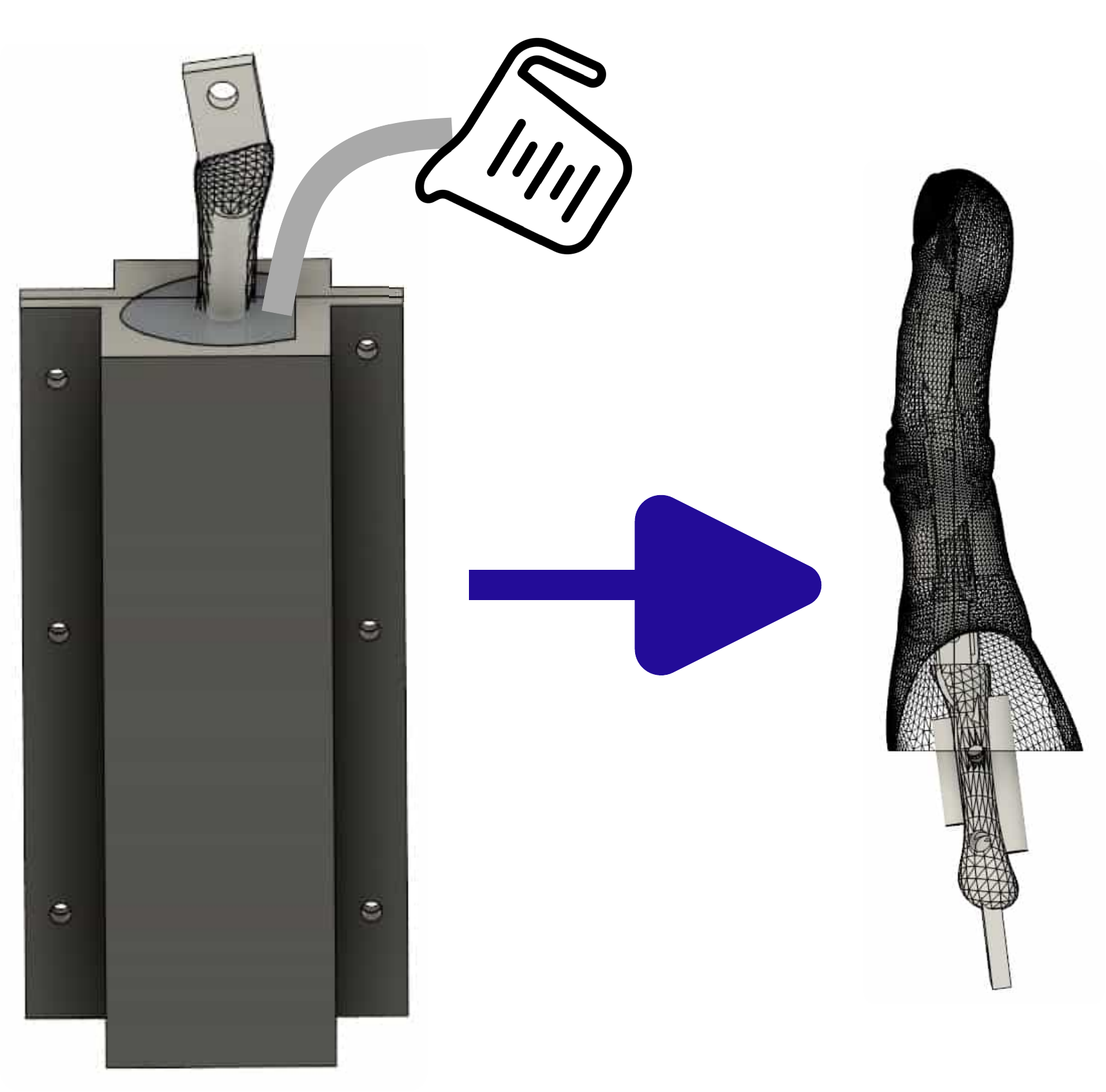}\label{fig:mold-step4}}
   \caption{(a) A top mold is designed to include outdents that match
     the finger skeleton shape. In the first step, connect this mold
     with the bottom mold. (b) Pour the silicone into the mold for
     4-hour curing, then remove the top mold and keep the cured
     silicone inside the bottom mold. (c) Place the bones with tendons
     into the created intends for the bones and screw on the other top
     mold. (d) Pour the silicone into the mold and, after 4-hour
     curing, the compliant skin surrounds the finger skeleton.}
   \label{fig:finger-fabrication}
 \end{figure*}

\section{DESIGN AND FABRICATION}
\label{sec:design}
Our proposed design is bio-inspired, compliant, low-cost, and
modular. The finger design integrates soft compliant skin that mimics
a human finger with an articulated rigid skeleton that provides
structural strength. The hand is low-cost and utilizes standard
components and digital manufacturing methods (e.g., 3D printing,
laser-cutting) with a simple and fast assembly process. Furthermore,
the hand design is modular --- fingers can be added and removed from
the design and can also be arranged in various configurations for
different scenarios.

\subsection{Cable-Driven Skeleton Design}
\label{sec:finger-design}

The key idea is to embed articulated rigid skeleton into soft
materials to form a controllable and sensible finger.  The design of
the robotic skeleton is shown in
Fig.~\ref{fig:index-finger-bones}. Each finger has four rigid bones
connected by the pin joints which allows us to easily measure the
joint angles. In every pin joint, we added a magnetic encoder
(Fig.~\ref{fig:bone-magnet}) and a diametrically magnetized magnet
(Fig.~\ref{fig:bone-sensor}) to measure the joint angle. This skeleton
structure is cable-driven. We added attachments to the front and the
back side of the bones for driving cables which are tied to the bones
through holes on their bottom and top sides
(Fig.~\ref{fig:bone-magnet}). Our skeleton design is bio-inspired. Our
bones are based on human finger bones, including the dimension and the
shape. Pin joints are used as ligaments and cables are used as finger
tendons. Currently two cables are attached to every finger --- one on
the front side fixed at the finger tip and the other one on the back
side fixed at one end of the proximal phalange shown in
Fig.~\ref{fig:index-finger-bones} and
Fig.~\ref{fig:skeleton-assembly}. One cable is used to bend the finger
forward and the other is used to bend it backwards. We can easily add
more tendons if necessary by adding holes to the bone and fixing more
cables.

\subsection{Fabrication Process}
\label{sec:fabrication}

We first assemble bones to form the finger skeleton shown in
Fig.~\ref{fig:skeleton-assembly}. All the bones are 3D printed.
Specifically, we use Nylon 12 (PA 2200) material (tensile strength
\SI{46}{MPa}, tensile modulus from \SIrange{1600}{1700}{MPa},
elongation at break 14--20\%). Magnets and angle sensor boards are
installed inside the bones. Each sensor board requires six wires which
are surrounded by a silicone tube. Then we connect all the bones with
pins and attach cables to the skeleton. Before molding the fingers, we
insert the cables through a silicone tube and then through the
attachments on the sides of the bones, to keep the cable secure and
away from silicone while molding.

We use a two step molding process to create the compliant skin for the
finger. The CAD models for the mold are created based on the finger
bones and the finger shape (Fig.~\ref{fig:finger-fabrication}). We 3D
print three mold pieces that are used to create a complete finger
using Onyx material (tensile modulus \SI{2.4}{GPa}, elongation at
break 25\%, tensile stress at yield \SI{40}{MPa}). We use room
temperature curable silicone (Smooth-On Ecoflex 00-20). In the first
step, we mold the bottom part of the finger using two mold pieces
(bottom and a top with outdents which match the bones) by pouring the
silicone into the mold shown in Fig.~\ref{fig:mold-step1} and
Fig.~\ref{fig:mold-step2}. After the curing (4 hours), the top part of
the mold is removed and the bones with tendons are placed into the
created intends for the bones (Fig.~\ref{fig:mold-step3}). The third
molding piece is screwed on and used for the top of the finger. After
filling the silicone into the new mold and curing for 4 hours, we
remove the hollow silicone tubes holding the cable and the finger is
complete (Fig.~\ref{fig:mold-step4}). Our finger design is
low-cost. All the components to fabricate a fully functional finger
are listed in Table~\ref{tab:cost}.

\begin{table}[t]
  \centering
  \caption{Cost of the Finger Components}
  \label{tab:cost}
  \begin{tabular}{l|cc}
    \toprule
    Material&Quantity&Price(\$)\\
    \midrule
    Silicone&\SI{15}{ml}&$\approx$2\\
    Bones&4&28\\
    Mold&3&35\\
    Angle Sensor Electronics&3&60\\
    Cables, Pins, Screws, Wires&&$\approx$1\\
    \midrule
    Total&&126\\
    \bottomrule
  \end{tabular}
\end{table}

\subsection{Mechanical Advantages}
\label{sec:advantage}

\begin{figure}[t!]
  \centering
  \subfloat[]{\includegraphics[width=0.18\textwidth]{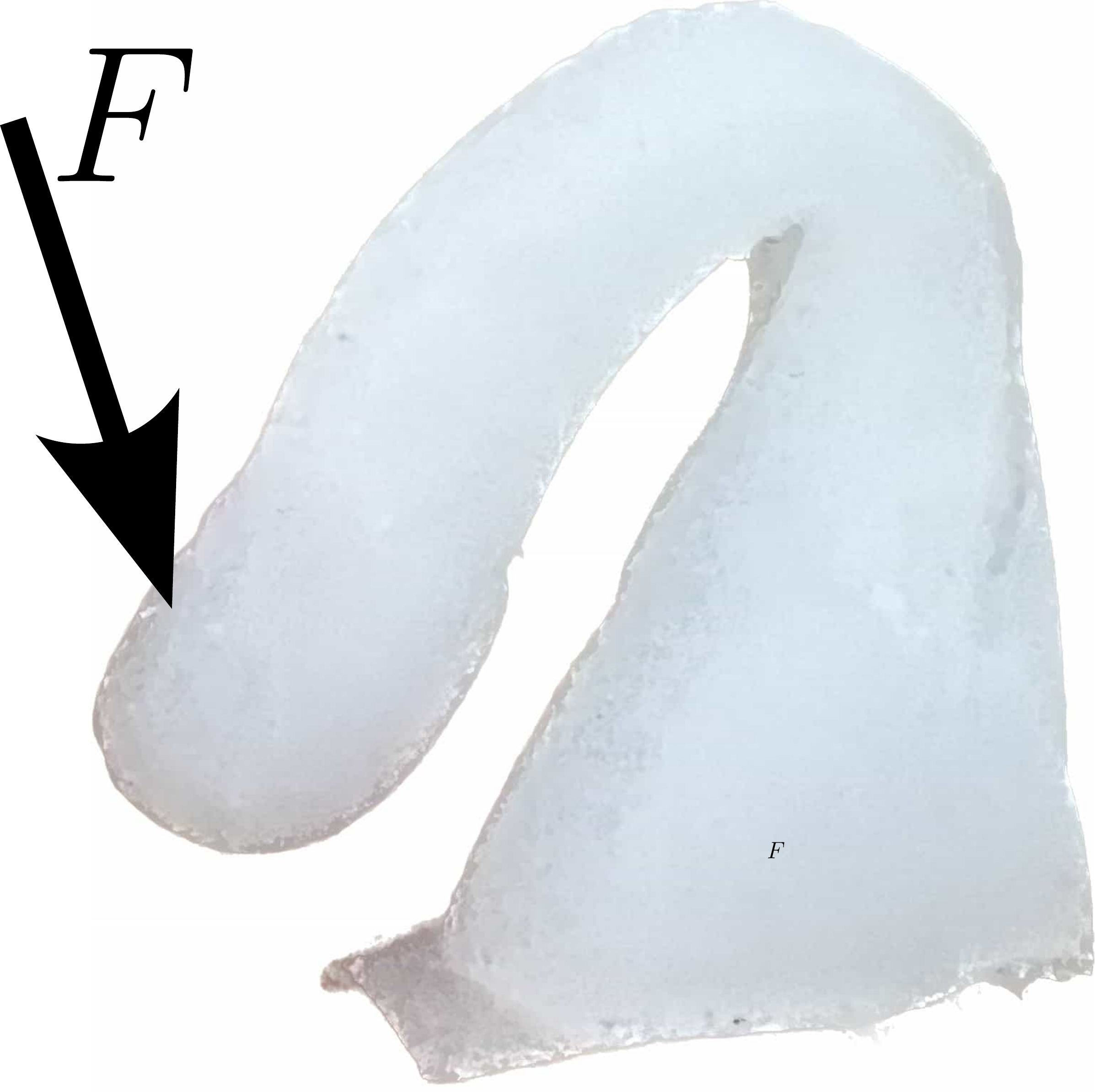}}
  \hfil
  \subfloat[]{\includegraphics[width=0.24\textwidth]{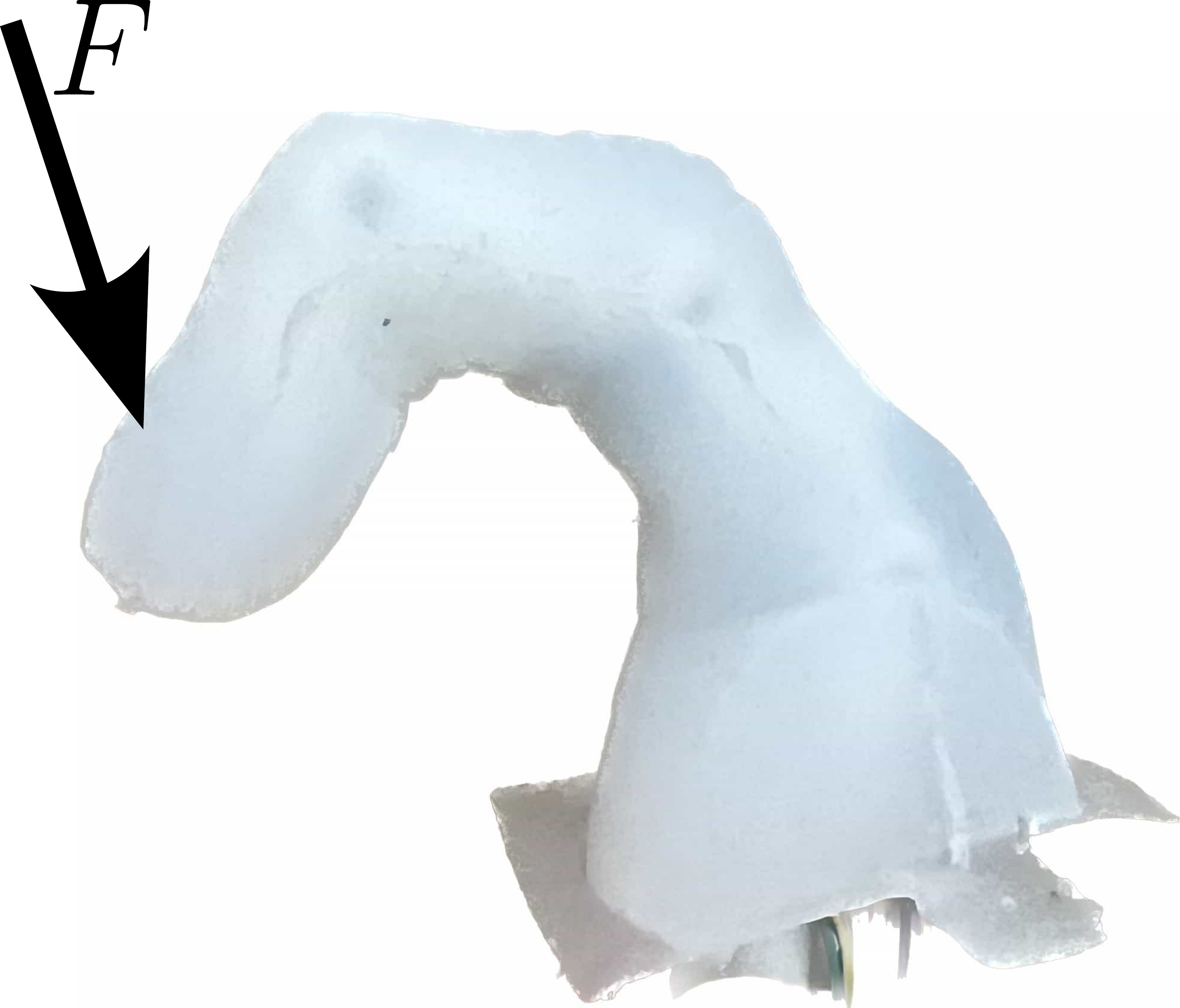}}
  \caption{(a) Bend a finger without rigid skeleton embedded. (b) Bend
  a finger with rigid skeleton embedded.}
  \label{fig:bend-finger}
\end{figure}

Embedding rigid skeleton structure into soft silicone can provide
mechanical advantages over pure rigid robots and soft robots. First,
the soft body is helpful to maintain the structure of the rigid
skeleton. The assembled rigid skeleton is delicate and cannot be
driven by cables without being surrounded by the soft body. The pin
joints can break easily under external load but the silicone can
tightly maintain the positions of all bones so that they can maintain
their pin joint connections well. Reversely, the rigid skeleton can
significantly increase the strength of the soft body made by silicone
because the bones are strongly connected via pin joints. In addition,
the rigid skeleton can define the shape of the soft body when
bending. The comparison between with and without the skeleton is shown
in Fig.~\ref{fig:bend-finger}. We applied forces to the tip of both
fingers, and the finger with the rigid skeleton is able to mimic the
human finger behavior well, namely the finger curves on the joint
locations and all the finger segments are straight. This mechanical
advantage is unique among previous soft finger designs and can be
helpful for human hand grasping and manipulation imitation.

\section{CONTROL AND POSE ESTIMATION}
\label{sec:control}

\subsection{Control Architecture}
\label{sec:architecture}

\begin{figure}[t]
  \centering
  \includegraphics[width=0.4\textwidth]{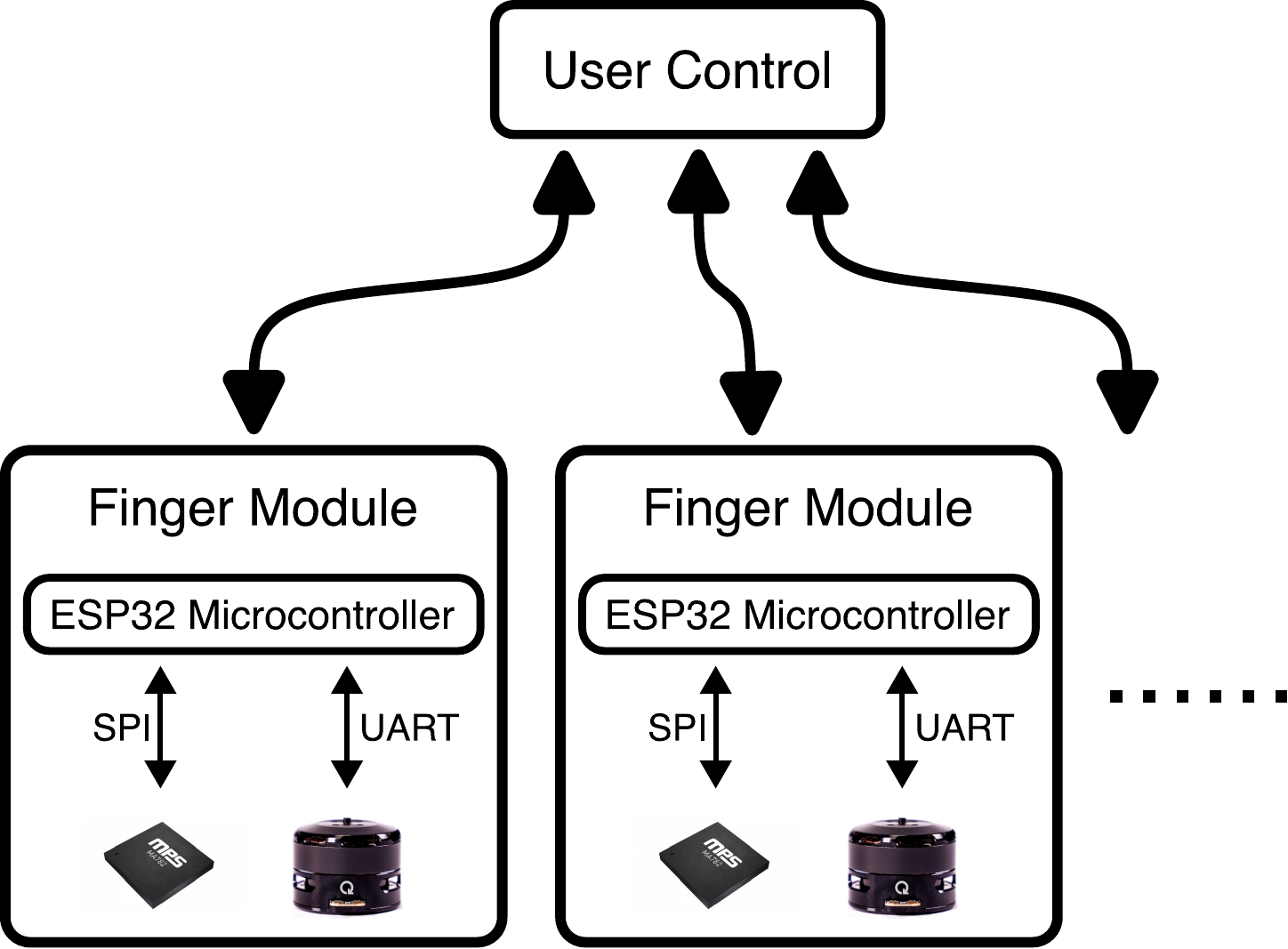}
  \caption{Control architecture of the modular hand system.}
  \label{fig:control-architecture}
\end{figure}

We design all fingers in a modular way --- every finger has its own
processor to handle low-level control and communication and can be
running independently. All the fingers are controlled by identical
driving systems and control boards. A central computer is
communicating with all fingers, including sending commands and
obtaining finger hardware state. The general control architecture is
shown in Fig.~\ref{fig:control-architecture}. This architecture makes
use of distributed computing power and also allows users to easily add
or remove finger modules.

\begin{figure}[t]
  \centering
  \includegraphics[width=0.3\textwidth]{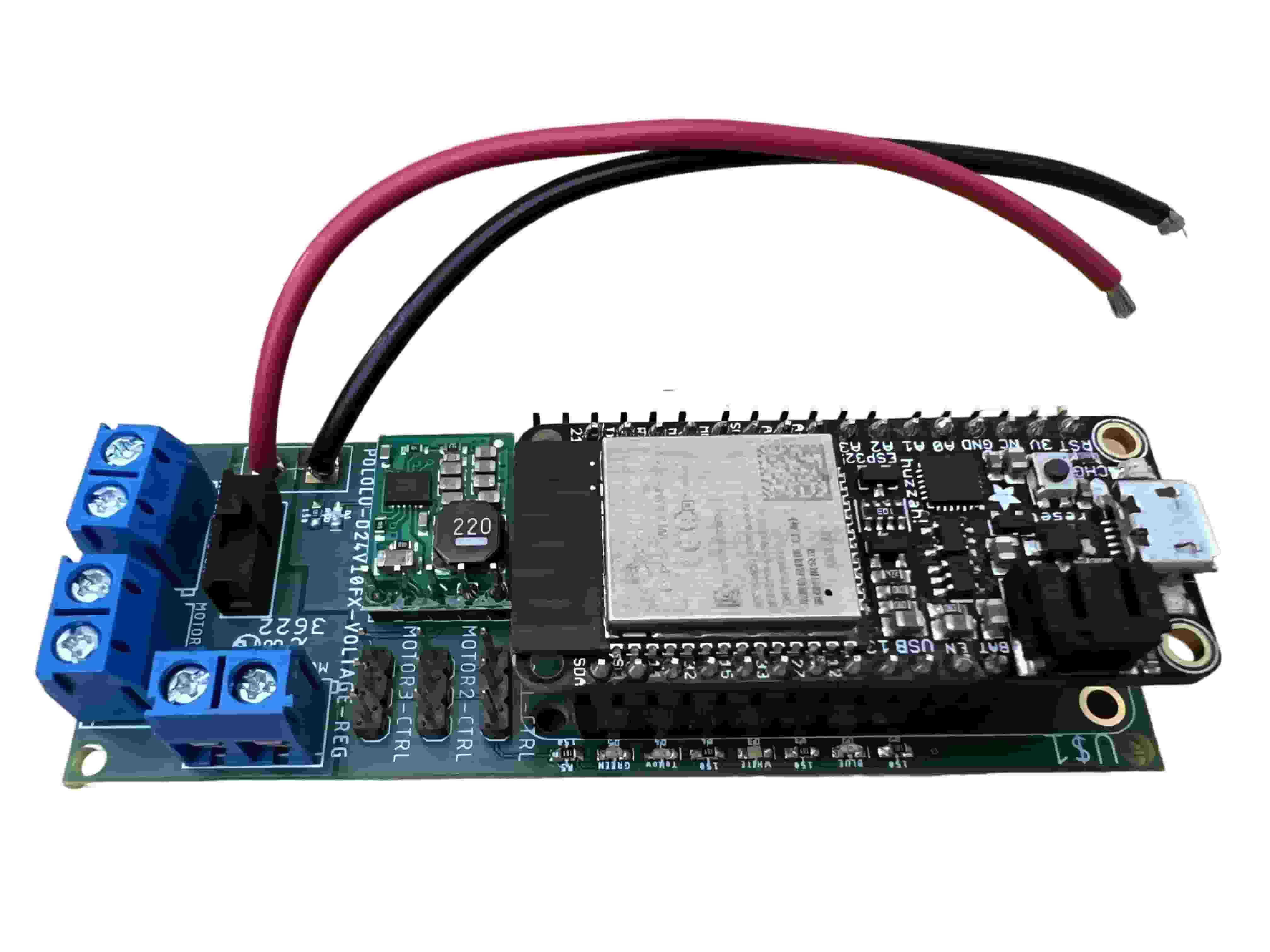}
  \caption{The electronics board for finger control. The board is able
  to control three brushless DC motors and communicate with three
  angle sensors.}
  \label{fig:control-board}
\end{figure}

The control board (Fig.~\ref{fig:control-board}) contains one
customized ESP32 microcontroller for communicating with three angle
sensors (MagAlpha MA782 from MonolithicPower, \SI{16}{Bit} resolution,
SPI interface) and controlling two brushless DC motors. The control
commands are sent by users. The pose of the finger is updated at
around \SI{200}{Hz} and the current angular positions of motors are
updated at around \SI{20}{Hz}. This real-time performance can be
useful for developing real-time grasping strategies. And the
resolution of the angle sensors enable a robotic finger to have high
sensitivity --- detecting tiny change of its shape and capturing
high-frequency tiny motions. This high sensitivity can be shown from
the experiment in Sec.~\ref{sec:pose-experiment}.

\subsection{Pose Estimation}
\label{sec:pose}

\begin{figure}[t]
  \centering
  \includegraphics[width=0.4\textwidth]{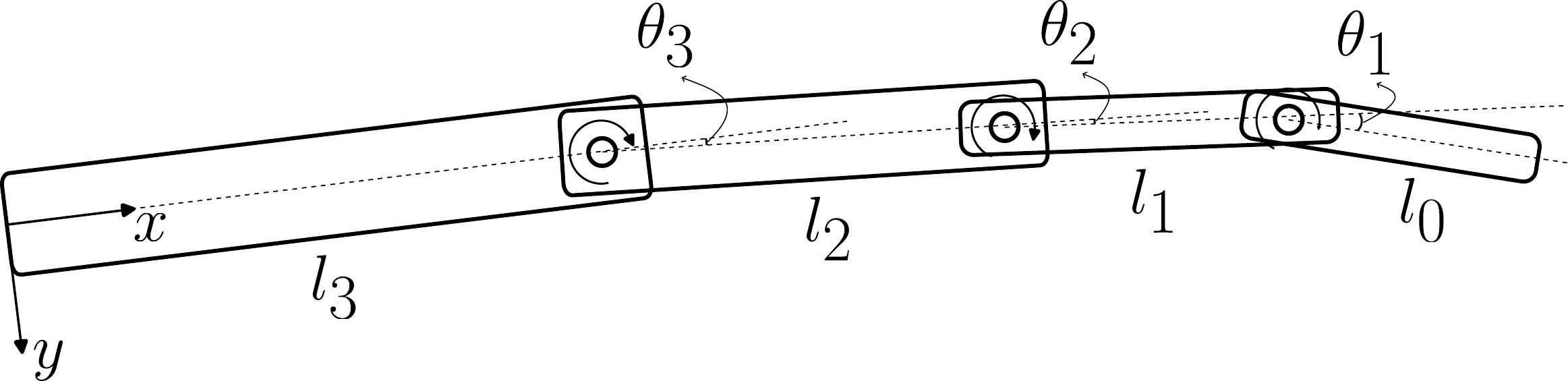}
  \caption{Kinematics model of the index finger skeleton.}
  \label{fig:skeleton-kinematics}
\end{figure}

The kinematics model of the robotic skeleton of the index finger is
shown Fig.~\ref{fig:skeleton-kinematics} and the state of the finger
can be fully defined by
$\left[\theta_1,\theta_2,\theta_3\right]^{\intercal}$ and the position
of the finger tip
$p_{\mathrm{tip}} = \left[ x_{\mathrm{tip}},y_{\mathrm{tip}}
\right]^{\intercal}$ with respect to the root of the finger can be
calculated by the following
\begin{equation}
  \label{eq:finger-tip}
  \left[
    \begin{array}{c}
      p_{\mathrm{tip}}\\
      1
    \end{array}
  \right] = H_3H_2H_1\left[
    \begin{array}{c}
      l_0\\0\\1
    \end{array}
  \right]
\end{equation}
in which $H_{i}=\left[
  \begin{array}{ccc}
    \cos\theta_i&-\sin\theta_i&l_i\\
    \sin\theta_i&\cos\theta_i&0\\
    0&0&1
  \end{array}
  \right]$.

\begin{figure}[t]
  \centering
  \subfloat[]{\includegraphics[height=0.25\textwidth]{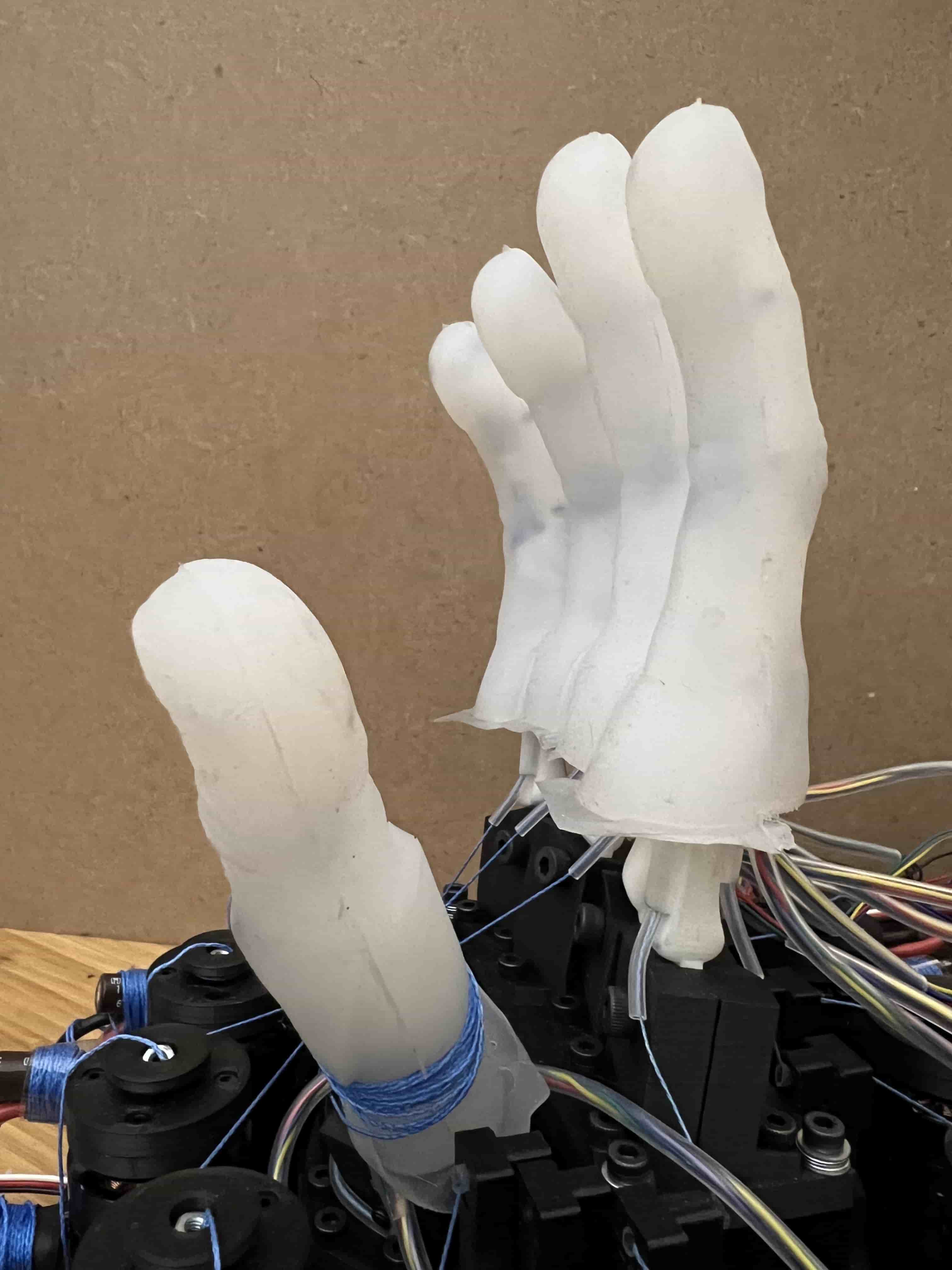}\label{fig:full-hand-setup}}
  \hfil
  \subfloat[]{\includegraphics[height=0.25\textwidth]{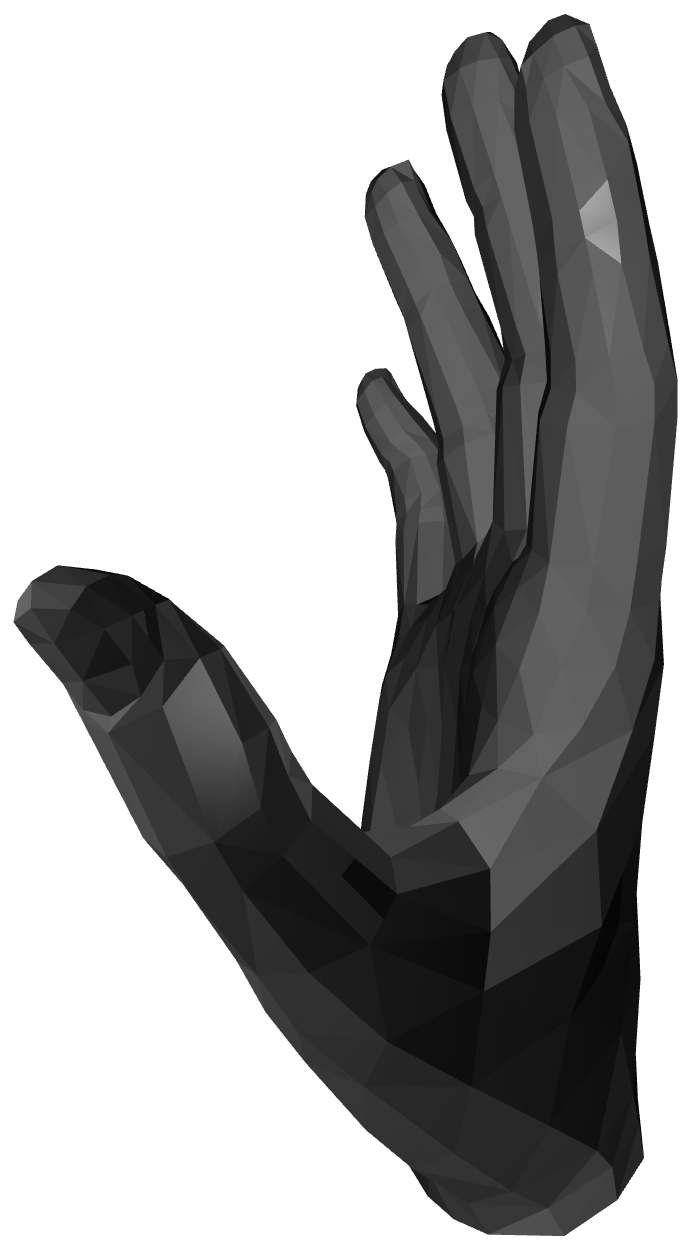}}
  \caption{A full hand (a) is visualized using MANO model (b).}
  \label{fig:finger-visualization}
\end{figure}

All finger poses are passed to a MANO model for real-time
visualization. In MANO model, each finger contains 3 joints and each
joint is considered as a ball joint defined by 3 parameters. In our
finger design, every joint is restricted to one DoF, so we just need
to update 3 parameters out of 9 for a single finger module. The
visualization result is shown in Fig.~\ref{fig:finger-visualization}.

\section{EXPERIMENTS}
\label{sec:experiment}

\subsection{Modular Design Versatility}
\label{sec:configuration}

\begin{figure}[t]
  \centering
  \subfloat[]{\includegraphics[width=0.2\textwidth]{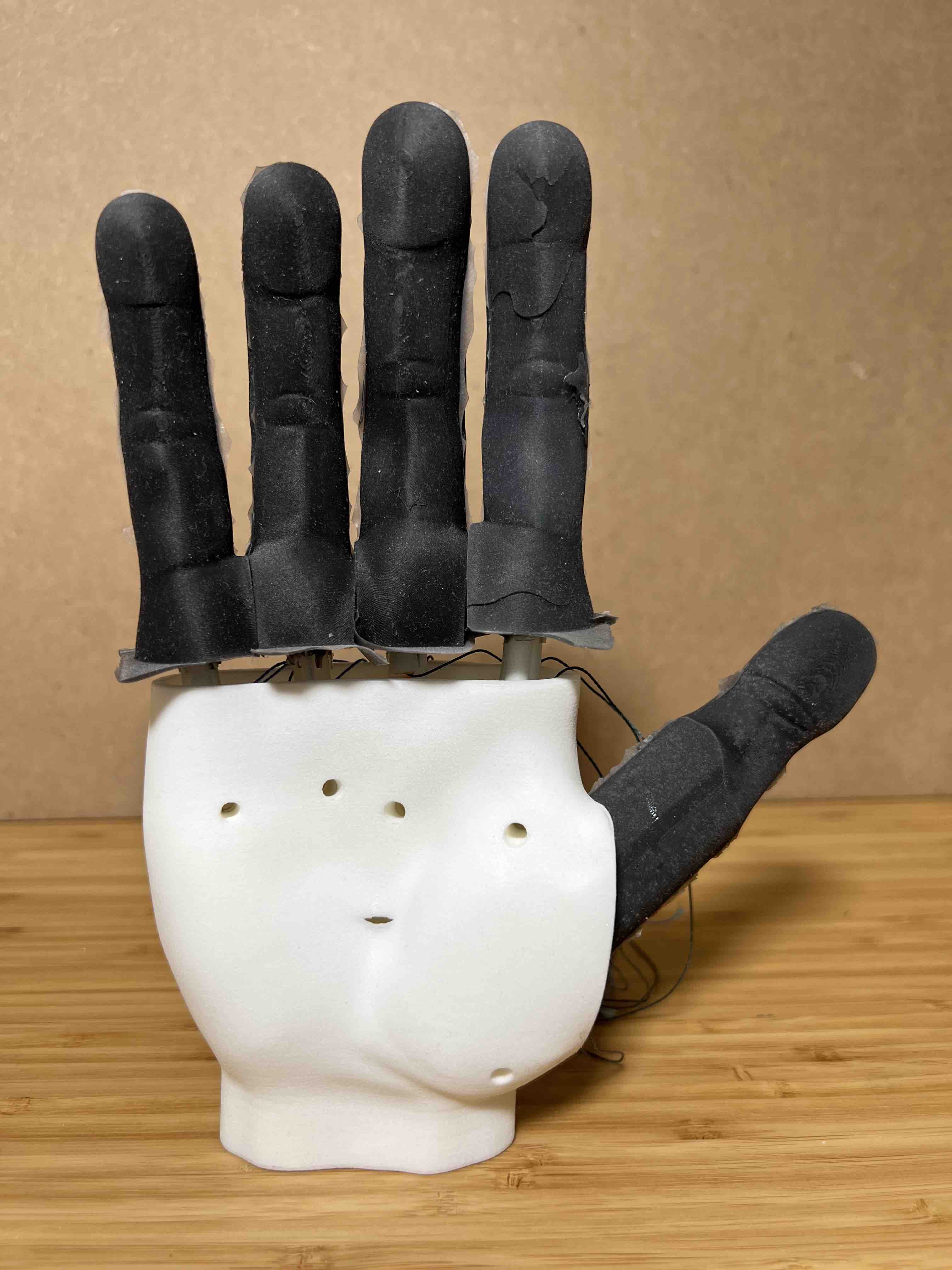}\label{fig:full-hand-config}}
  \hfil
  \subfloat[]{\includegraphics[width=0.2\textwidth]{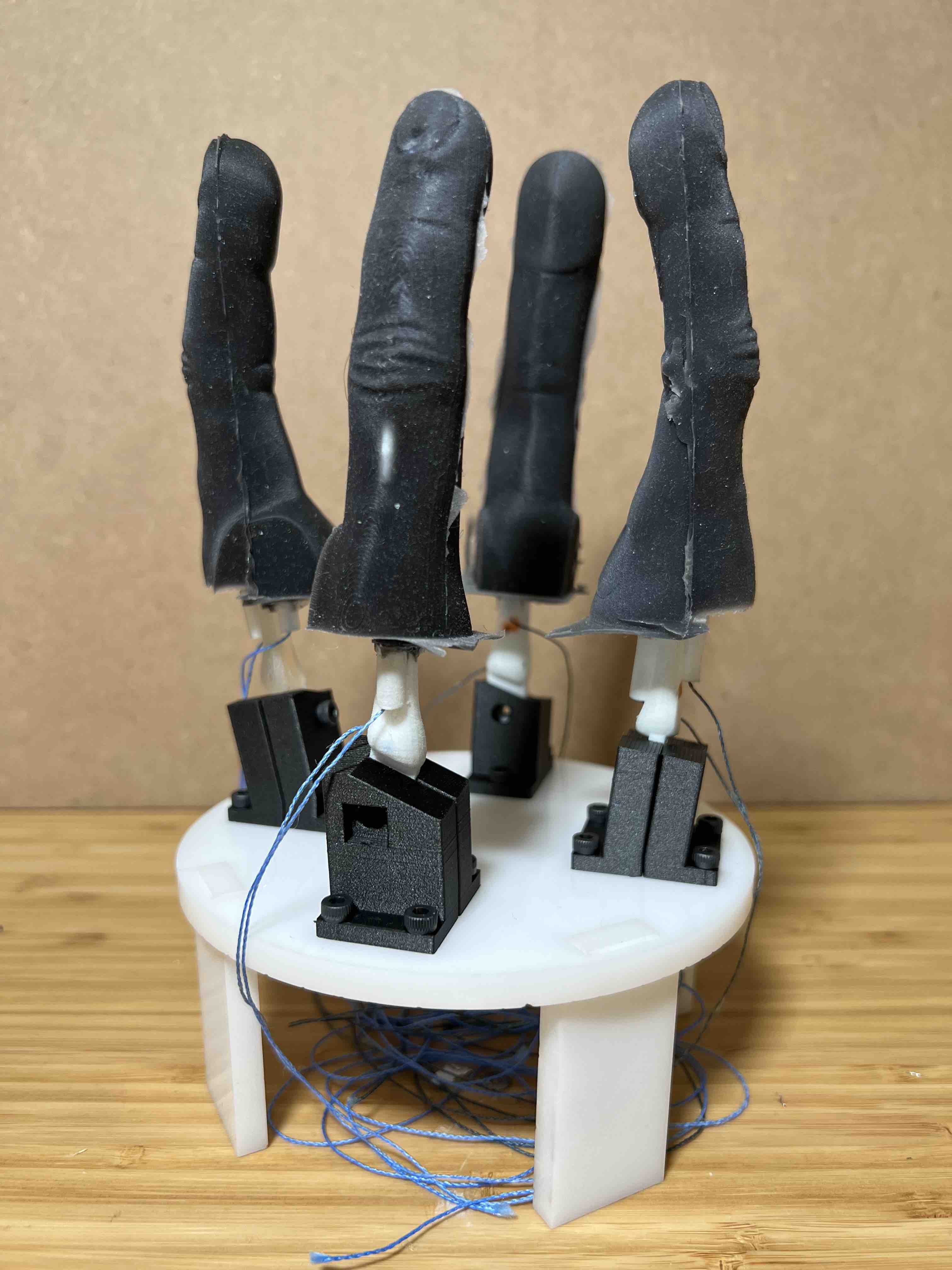}\label{fig:four-finger-config}}
  \caption{(a) A full hand configuration with five fingers mounted on
    a palm. (b) A full hand configuration with five fingers mounted on
    an acrylic base.}
  \label{fig:full-hand}
\end{figure}

The modularity of our finger design allows us to easily build various
configurations using our fingers that is similar to modular robots
since the fingers can be rearranged into different morphologies
similar to~\cite{Liu-smores-reconfig-ral-2019}. We fabricated all five
fingers (index finger, middle finger, ring finger, little finger,
thumb) and first assembled them in a human hand configuration shown in
Fig.~\ref{fig:full-hand-config}. We can also easily mount four fingers
on a laser-cut acrylic base shown in
Fig.~\ref{fig:four-finger-config}.

\subsection{Accurate Pose Estimation and High Sensitivity}
\label{sec:pose-experiment}

\begin{figure}[b!]
  \centering
  \subfloat[]{\includegraphics[height=0.25\textwidth]{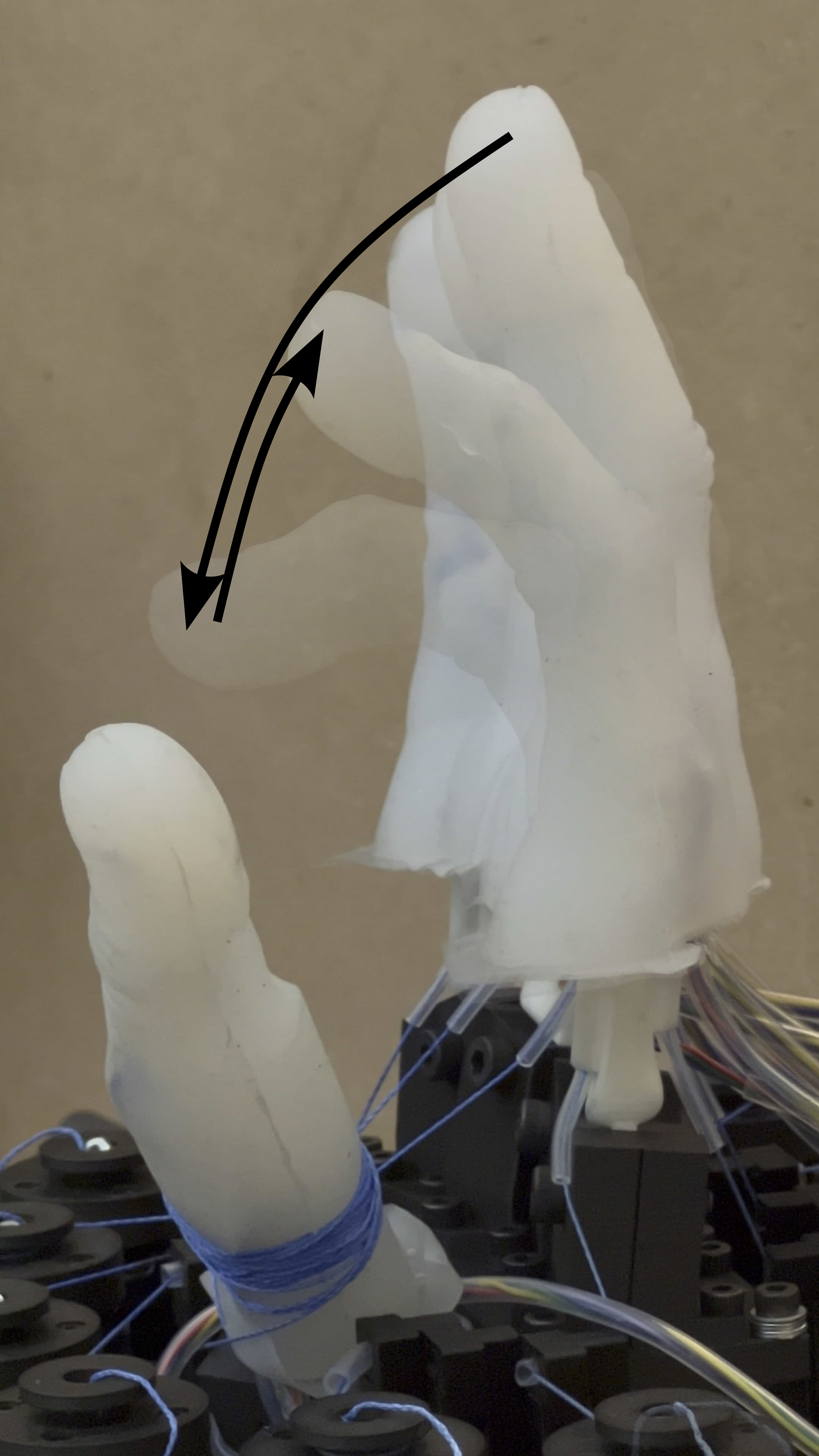}\label{fig:finger-motion-test-real}}
  \hfil
  \subfloat[]{\includegraphics[height=0.23\textwidth]{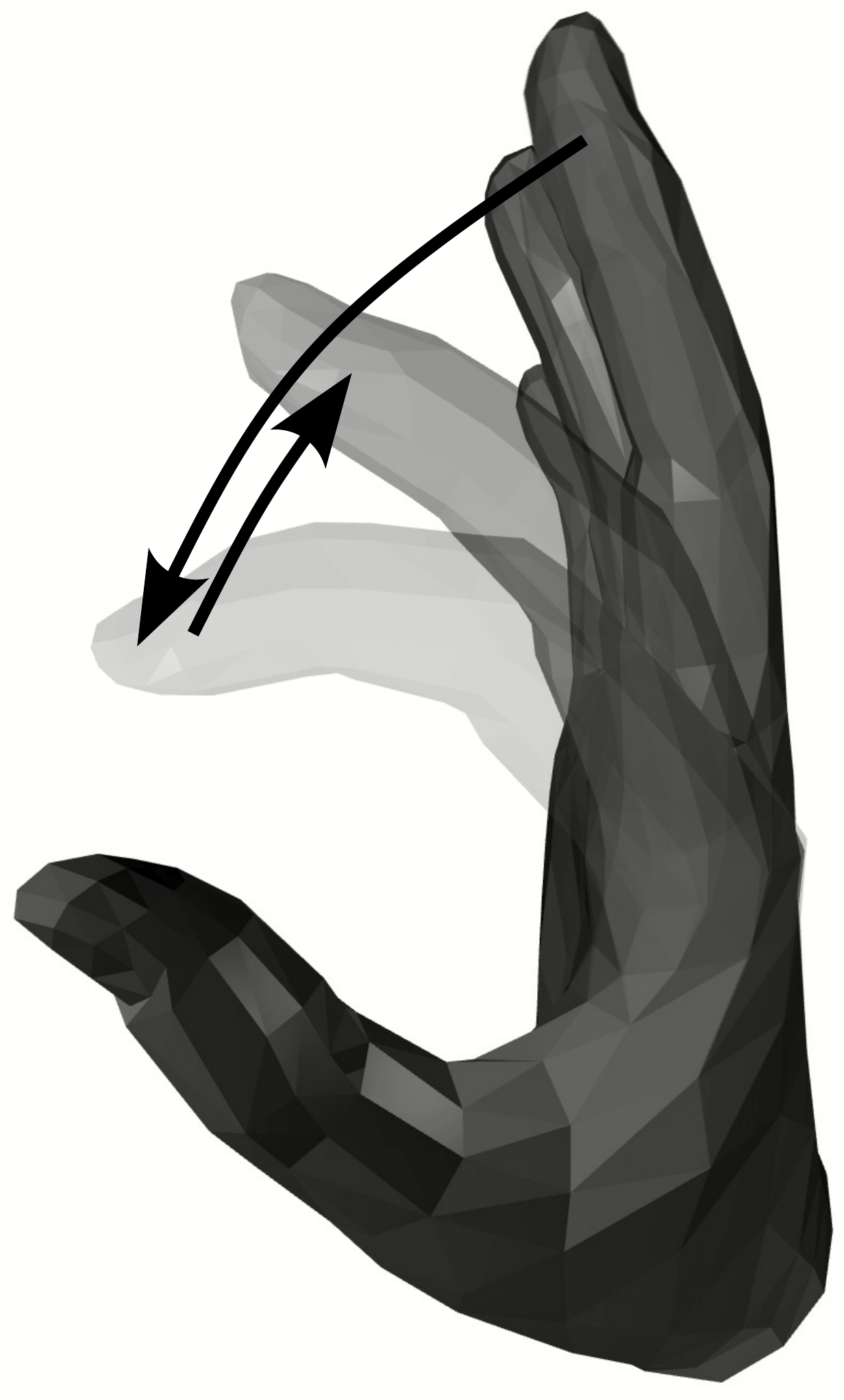}\label{fig:finger-motion-test-mano}}
  \caption{(a) Command the index finger to first bend and then
    extend. (b) Visualization result of MANO model.}
  \label{fig:finger-motion-test}
\end{figure}

\begin{figure}[t]
  \centering
  \subfloat[]{\includegraphics[width=0.24\textwidth]{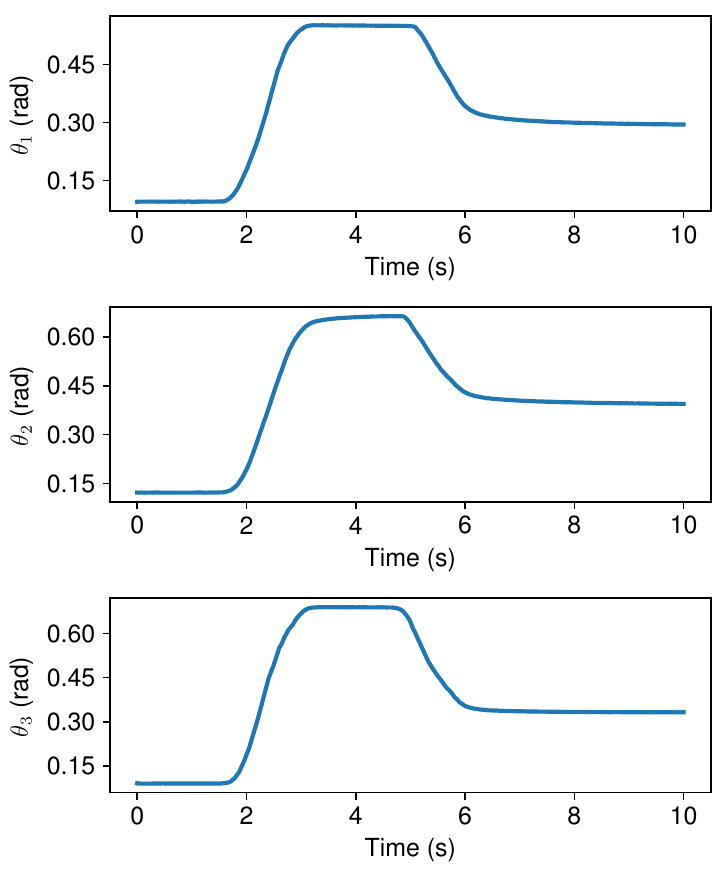}\label{fig:finger-motion-measure}}
  \hfil
  \subfloat[]{\includegraphics[width=0.24\textwidth]{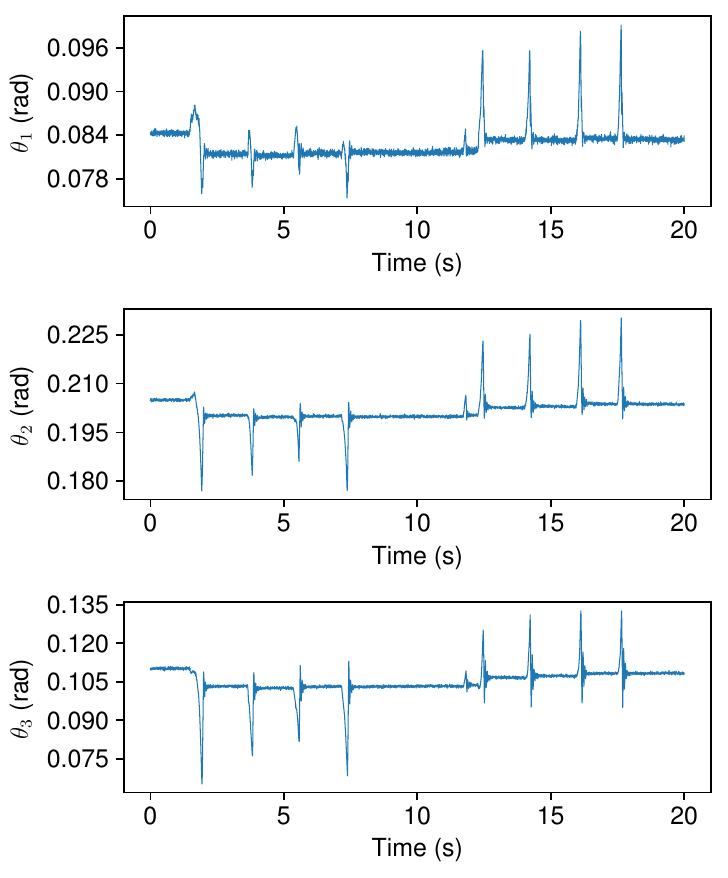}\label{fig:finger-shake-measure}}
  \caption{(a) Angle sensor data during the motion of the finger. (b)
    Angle sensor data while the finger tip being touched.}
\end{figure}

Although the rigid skeleton is fragile, surrounding it by silicone is
able to provide precise and real-time pose estimation capability. We
commanded the index finger to first bend and then extend to an
intermediate pose. The angle sensors can estimate the finger pose in
real time: the MANO model (Fig.~\ref{fig:finger-motion-test-mano}) and
the real hardware (Fig.~\ref{fig:finger-motion-test-real}) matched
well, and the sensor measurement is shown in
Fig.~\ref{fig:finger-motion-measure}.

\begin{figure}[t]
  \centering
  \includegraphics[width=0.2\textwidth]{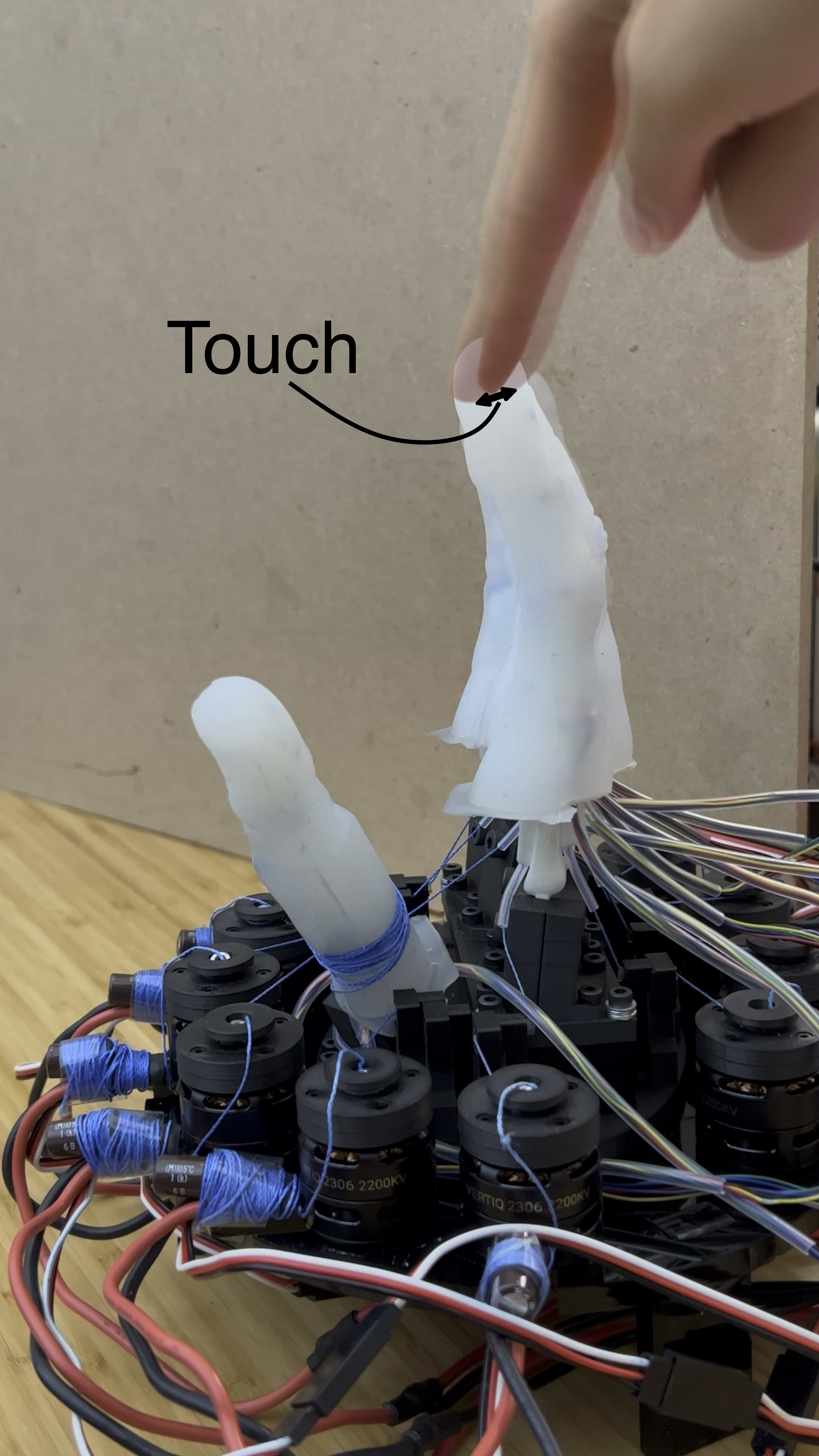}
  \caption{A person slightly touched the index finger tip multiple times.}
  \label{fig:finger-shake-real}
\end{figure}

Our finger can further detect tiny external loads, such as human
touch. In this experiment, a person slightly touched the robotic index
finger tip multiple times as shown in
Fig.~\ref{fig:finger-shake-real}. The quick and tiny change of the
finger shape can be clearly captured by the finger sensing capability
shown in Fig.~\ref{fig:finger-shake-measure}.

\begin{figure*}[t]
  \centering
  \subfloat[]{\includegraphics[width=0.2\textwidth]{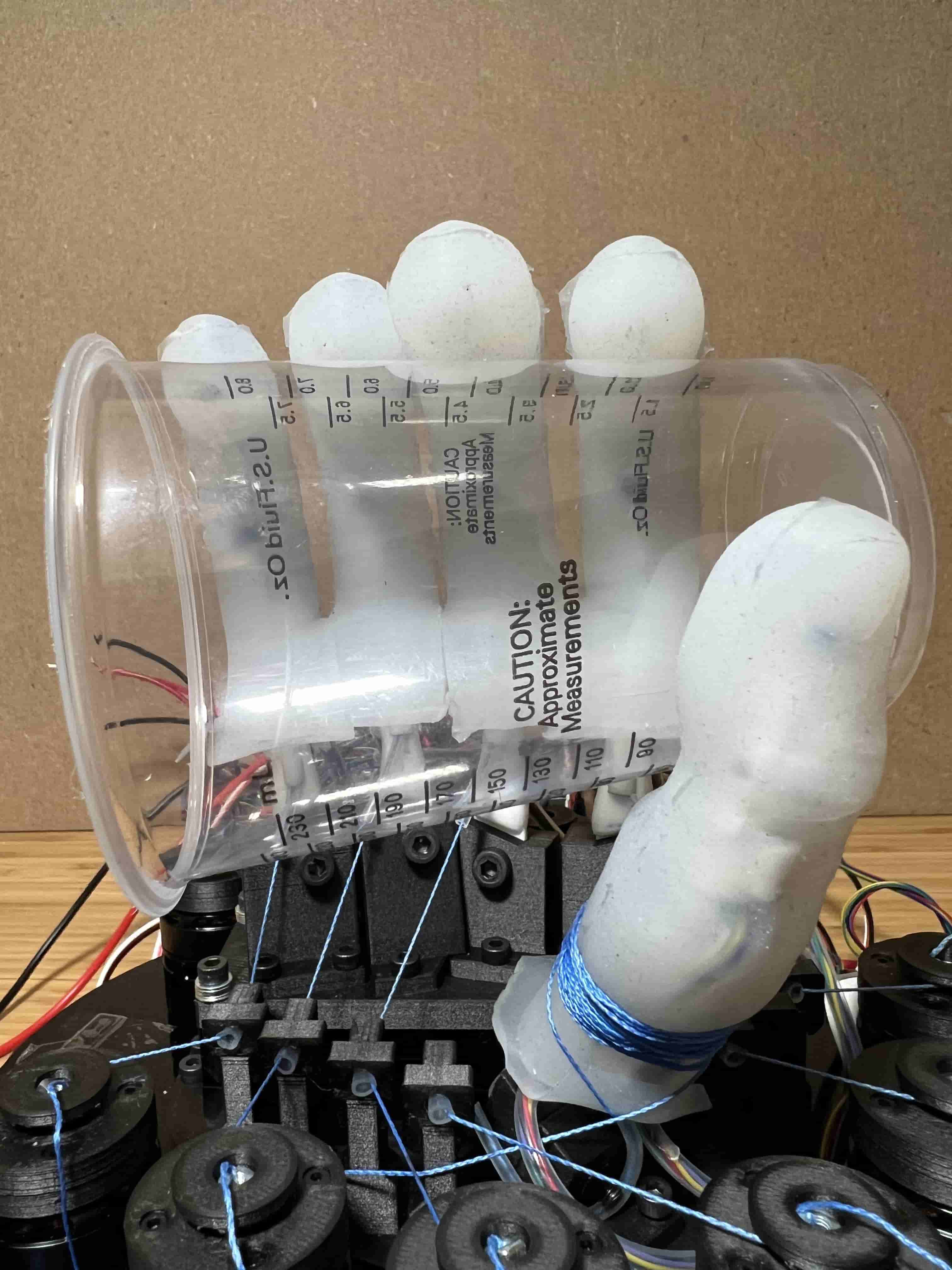}}
  \hfill
  \subfloat[]{\includegraphics[width=0.2\textwidth]{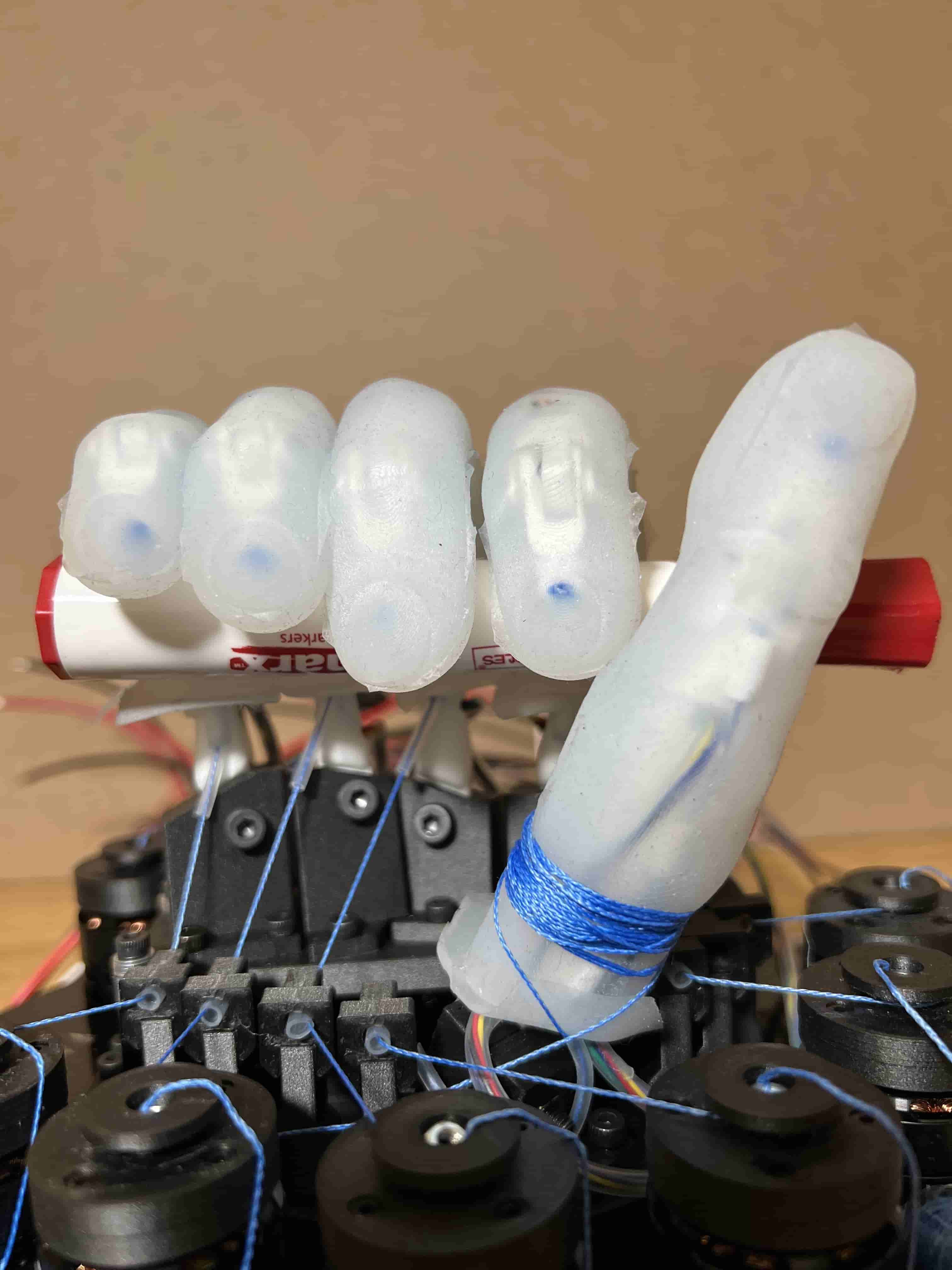}}
  \hfill
  \subfloat[]{\includegraphics[width=0.2\textwidth]{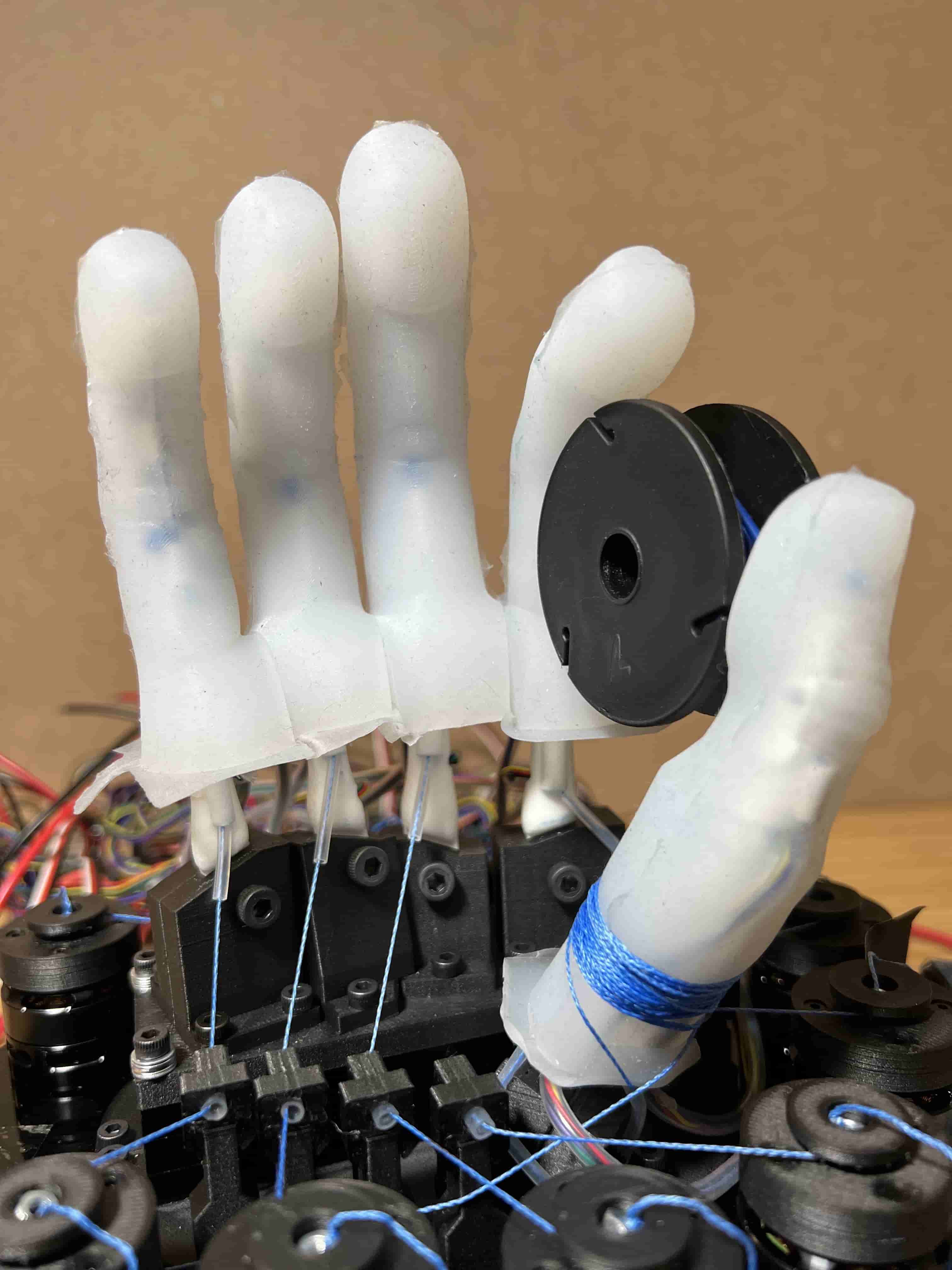}}
  \hfill
  \subfloat[]{\includegraphics[width=0.2\textwidth]{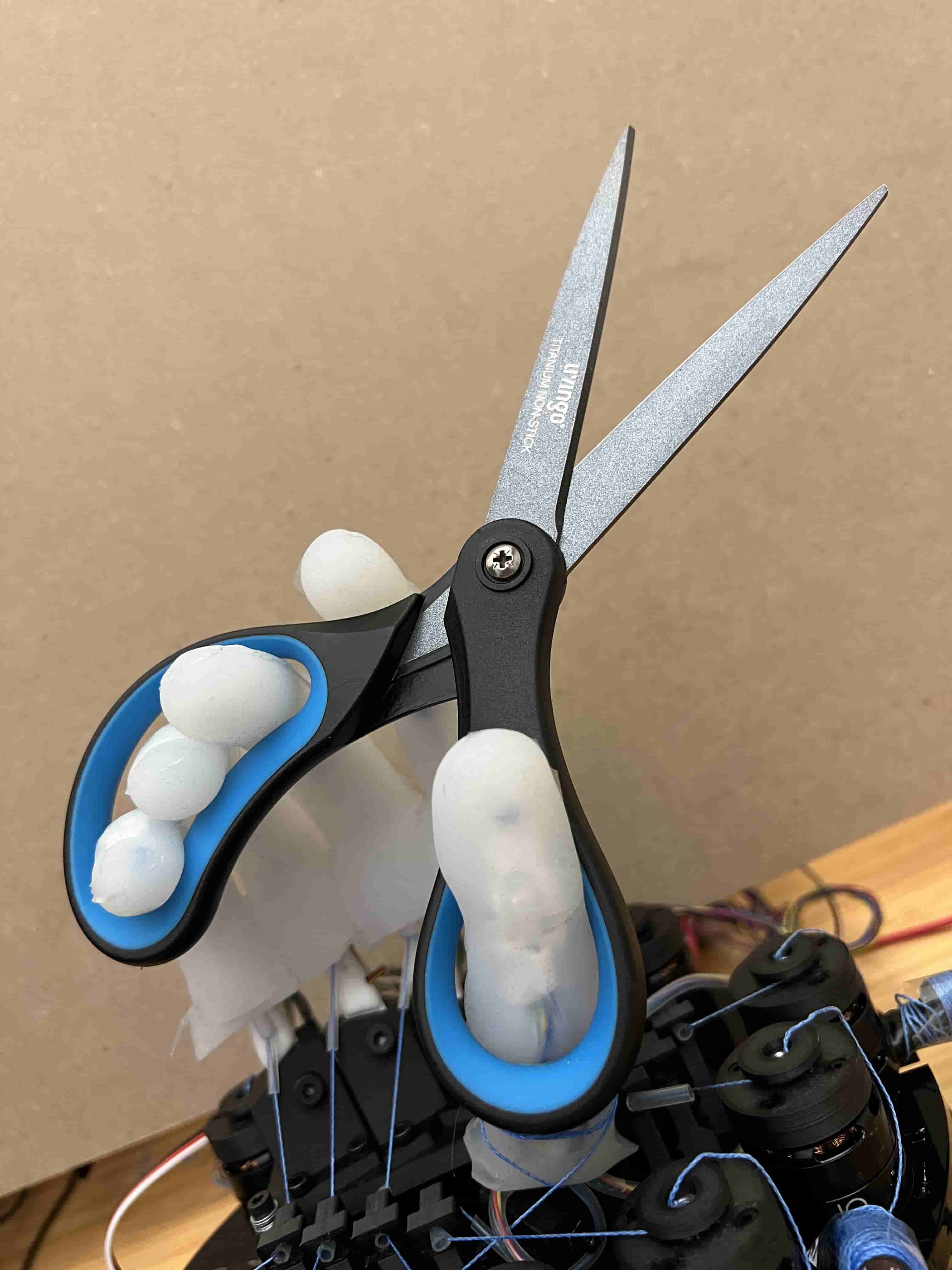}}\\
  \subfloat[]{\includegraphics[width=0.2\textwidth]{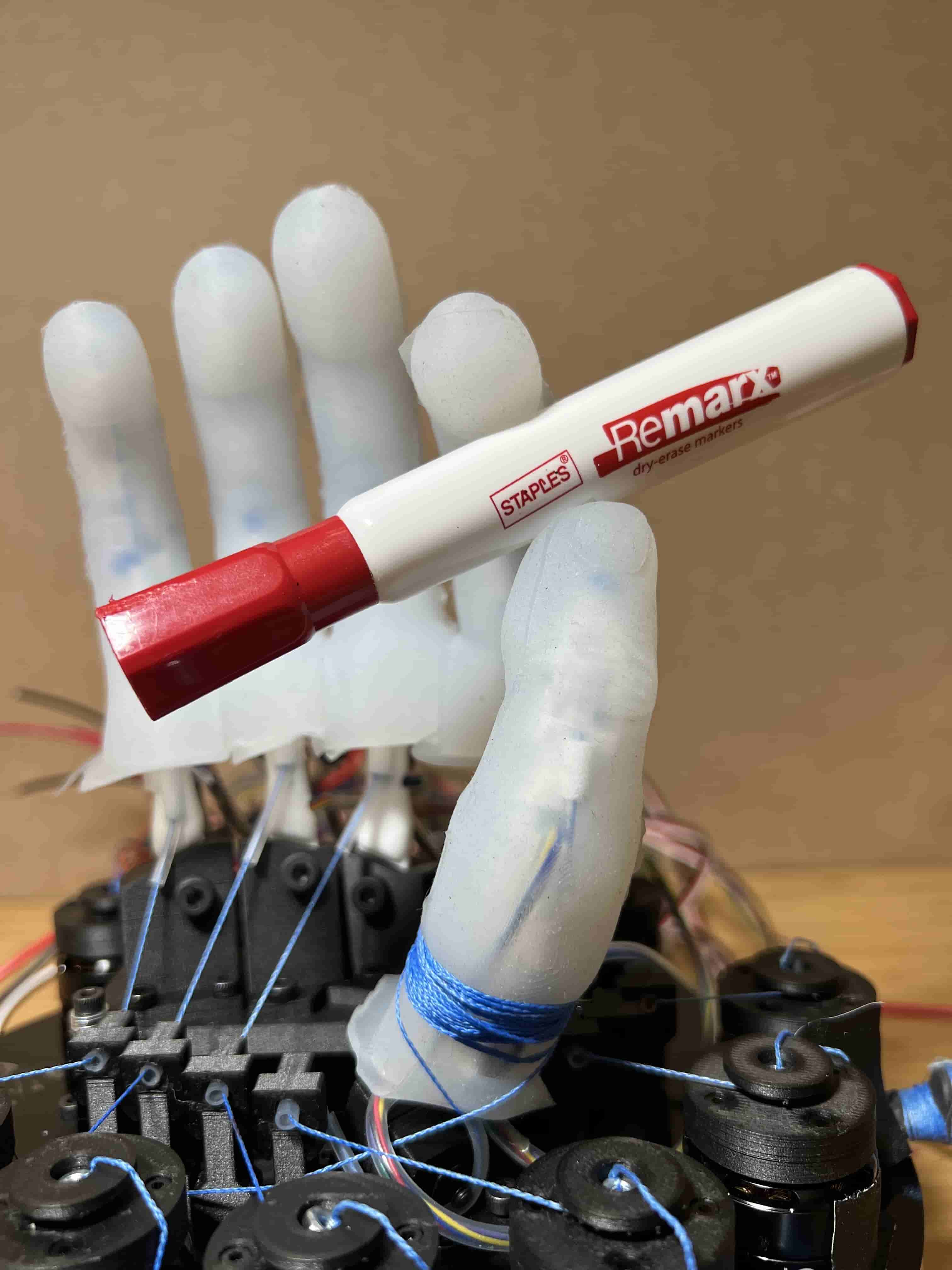}}
  \hfill
  \subfloat[]{\includegraphics[width=0.2\textwidth]{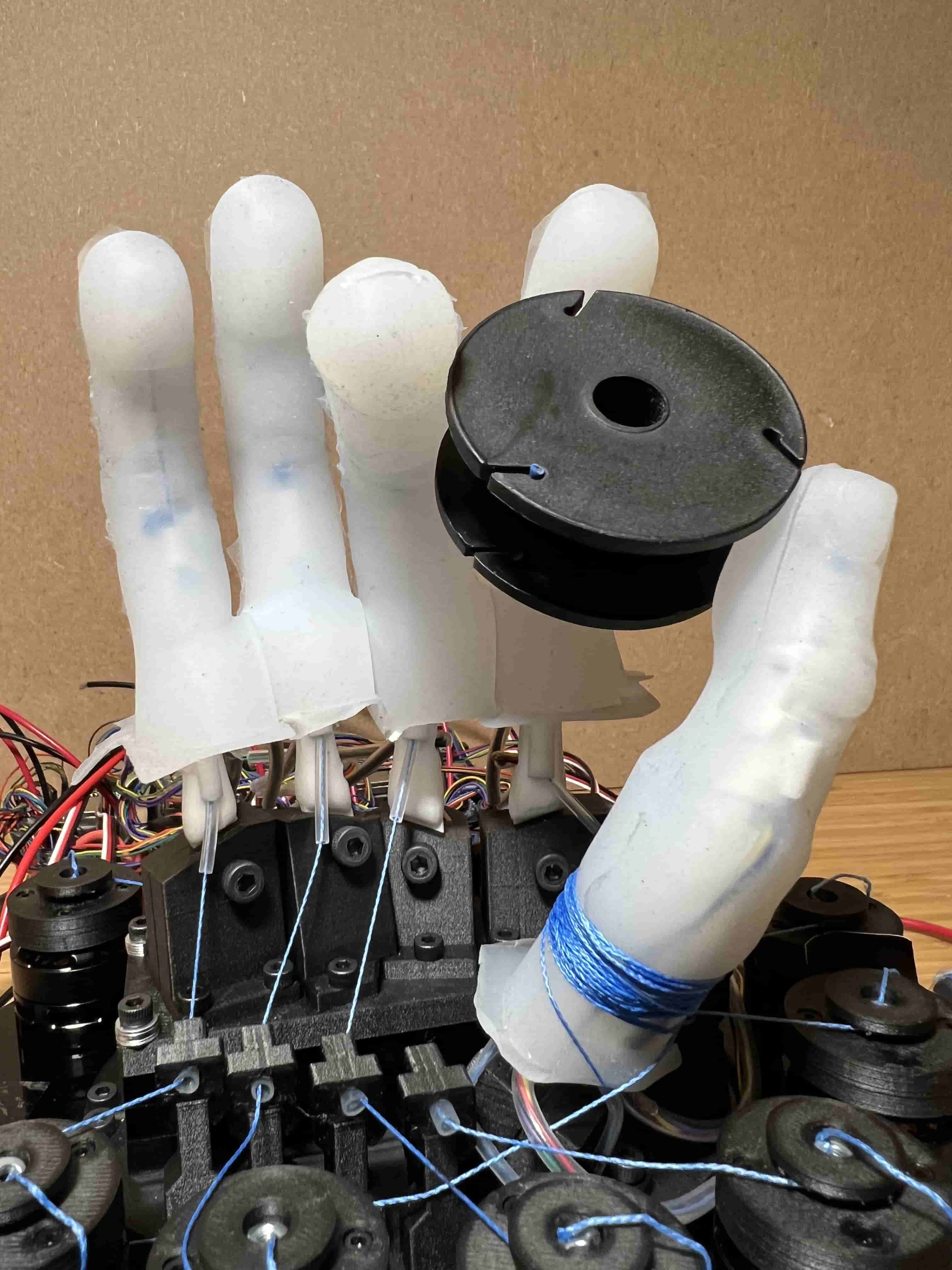}}
  \hfill
  \subfloat[]{\includegraphics[width=0.2\textwidth]{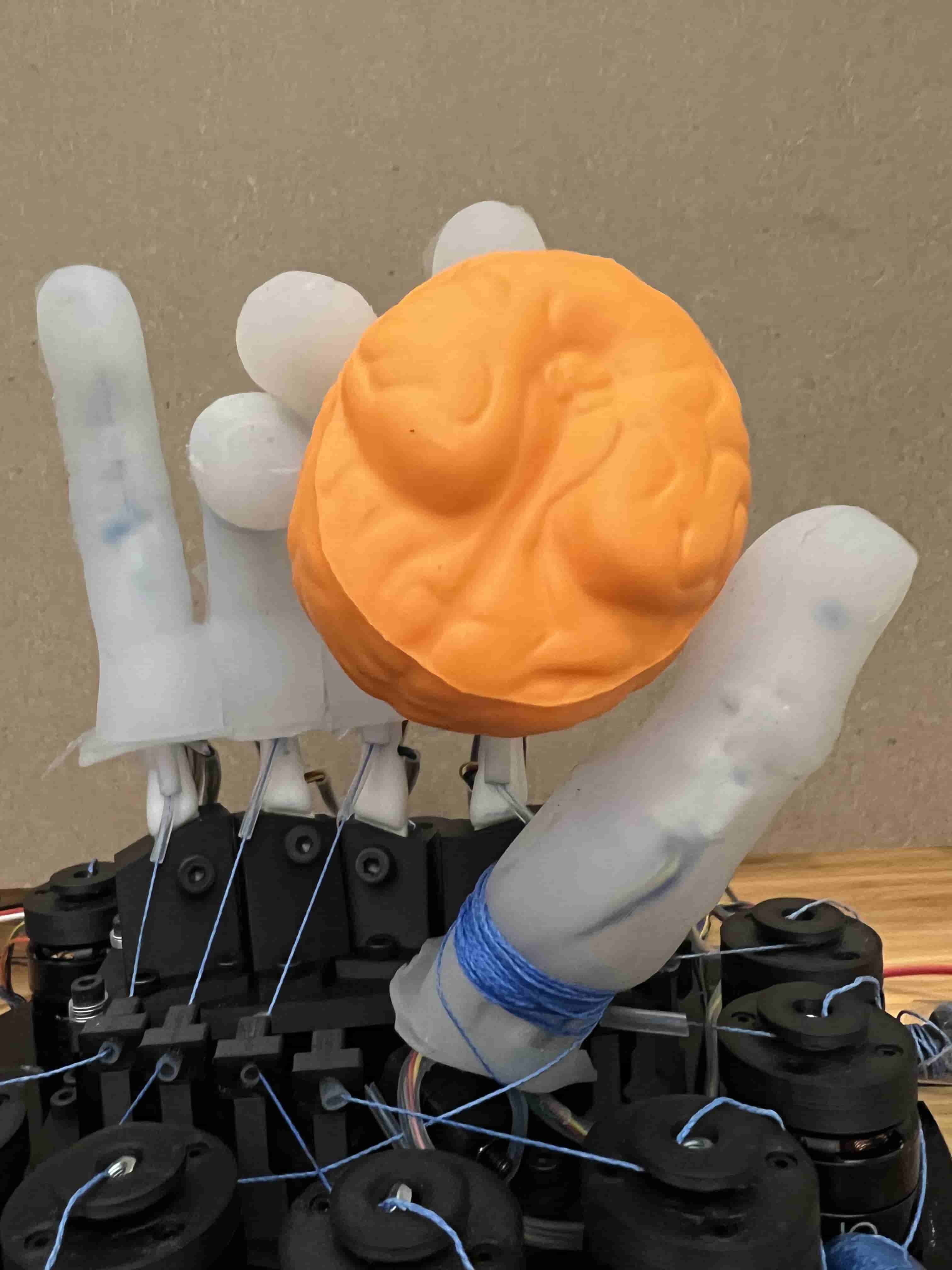}}
  \hfill
  \subfloat{\includegraphics[width=0.2\textwidth]{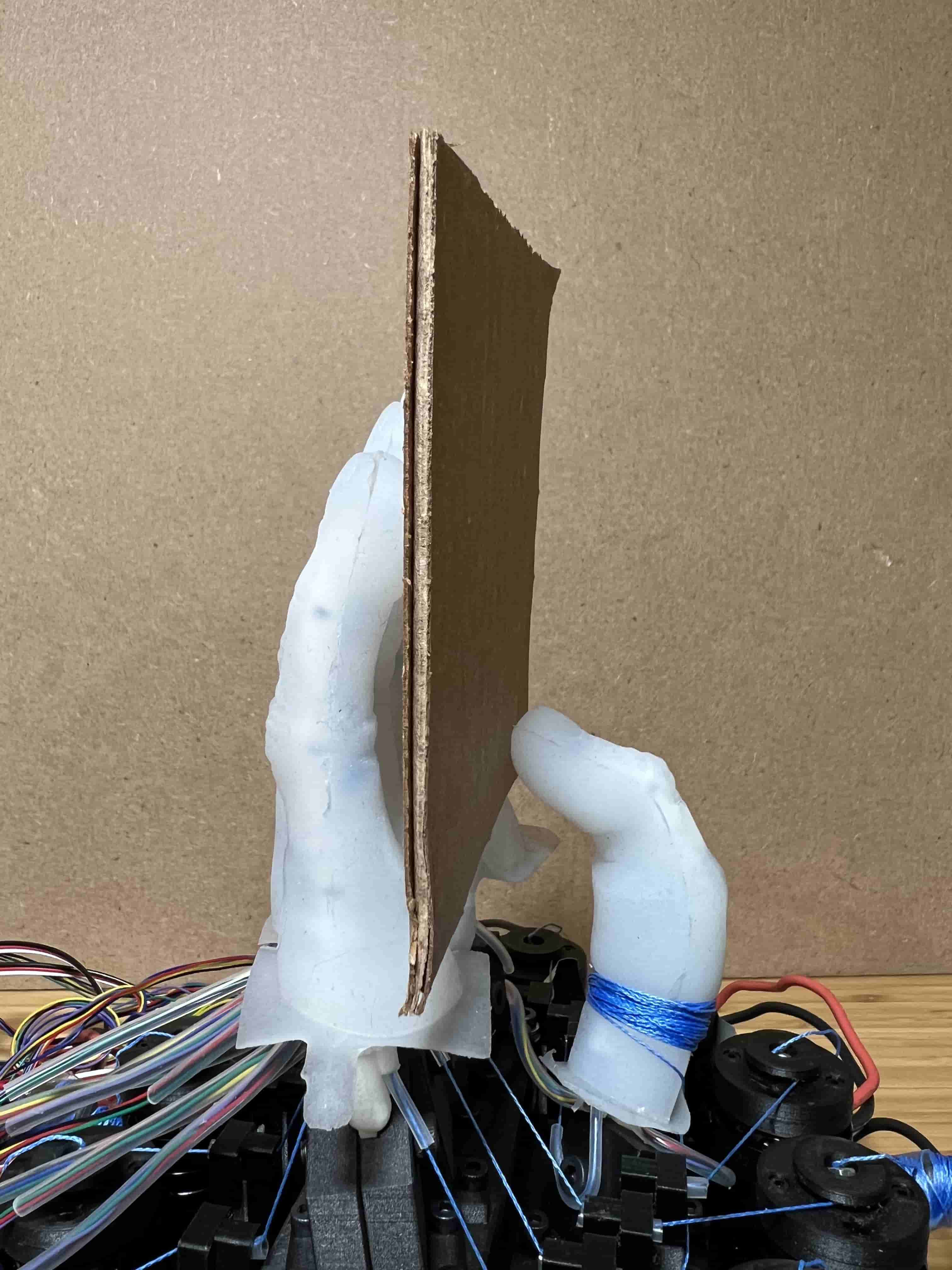}}
  \caption{(a) Large diameter. (b) Small diameter. (c) Ring. (d) Distal. (e) Tip
    pinch. (f) Tripod. (g) Quadpod. (h) Parallel extension.}
  \label{fig:grasp-types}
\end{figure*}

\subsection{Grasping Taxonomy}
\label{sec:grasp-type}

Human grasp types are synthesized into a grasp taxonomy that contains
33 different types~\cite{Feix-human-grasp-taxonomy-2015}. We set up
five fingers shown in Fig.~\ref{fig:full-hand-setup}. In this setup,
compared with a human hand, the palm is missing and the thumb is able
to bend but not capable of rotating around. Hence, this robotic hand
cannot handle grasp types that require the palm and different
positions of the thumb with respect to the other four fingers. Some
grasp types are shown in Fig.~\ref{fig:grasp-types}.

\section{CONCLUSIONS}

In this paper, we present soft modular robotic fingers. Our design is
bio-inspired --- embedding rigid skeletons into soft tissues. We
cloned and modified the human finger bones to create robotic finger
skeletons with high resolution angle sensors embedded. A carefully
designed fabrication process is introduced to quickly mold silicone
around rigid robotic skeletons to form human-like fingers. The
combination of the rigid elements with the soft materials is able to
create a robotic finger with real-time precise pose estimation and
high sensitivity. We further constructed a robotic hand with five
fingers and demonstrated its grasping capability by performing several
human grasp types. Future work includes a palm design with an
additional rotation DoF for the thumb, a compact design of the wrist
to hold the electronics, and the development of the control strategy
for dexterous manipulation and grasping.

\addtolength{\textheight}{-10.8cm}
%\addtolength{\textheight}{-12cm}   % This command serves to balance the column lengths
                                  % on the last page of the document manually. It shortens
                                  % the textheight of the last page by a suitablees amount.
                                  % This command does not take effect until the next page
                                  % so it should come on the page before the last. Make
                                  % sure that you do not shorten the textheight too much.

%%%%%%%%%%%%%%%%%%%%%%%%%%%%%%%%%%%%%%%%%%%%%%%%%%%%%%%%%%%%%%%%%%%%%%%%%%%%%%%%

\bibliographystyle{IEEEtran}
\bibliography{reference.bib}

\end{document}